\documentclass{article}

\PassOptionsToPackage{numbers, compress}{natbib}
\usepackage[preprint]{neurips_2025}

\usepackage{url}            
\usepackage{booktabs}      
\usepackage{amsfonts}       
\usepackage{nicefrac}      
\usepackage{microtype}
\usepackage[dvipsnames]{xcolor}
\usepackage{enumitem}
\usepackage{pifont}

\usepackage{graphicx}
\usepackage{subfigure}
\usepackage{multirow}
\usepackage{multicol}
\usepackage{amsmath}
\usepackage{amssymb}
\usepackage{mathtools}
\usepackage{bm}
\usepackage{amsthm}
\usepackage{algorithmic}
\usepackage{algorithm}
\usepackage{tikz}
\usepackage{wrapfig}
\usepackage{caption}
\usepackage{booktabs}
\usepackage{multirow}
\usepackage{graphicx}
\usepackage{colortbl}
\usepackage[normalem]{ulem}
\useunder{\uline}{\ul}{}

\newcommand{\name}{DiffE2E}
\definecolor{mine}{RGB}{205, 232, 248}
\definecolor{customblue}{HTML}{8FAADC}

\usepackage{booktabs} 
\usepackage{array}
\usepackage{multirow}
\usepackage{multicol}
\usepackage{color, colortbl}
\usepackage{amsmath}
\usepackage{amssymb}
\usepackage{hyperref}
\usepackage{cleveref}

\title{DiffE2E: Rethinking End-to-End Driving with a Hybrid Action Diffusion and Supervised Policy}

\author{%
    Rui Zhao$^1$\thanks{Contact me at \texttt{rzhao@jlu.edu.cn}} ~~Yuze Fan$^1$ ~~Ziguo Chen$^1$ ~~Fei Gao$^{1,2}$\thanks{Correspondence to:
    Fei Gao (\texttt{gaofei123284123@jlu.edu.cn})} ~~Zhenhai Gao$^{1,2}$\\
    {\small$^1$College of Automotive Engineering, Jilin University} \\ 
    {\small$^2$National Key Laboratory of Automotive Chassis Integration and Bionics, Jilin University} \\
}

\begin{document}

\maketitle

\begin{abstract}
    End-to-end learning has emerged as a transformative paradigm in autonomous driving. However, the inherently multimodal nature of driving behaviors and the generalization challenges in long-tail scenarios remain critical obstacles to robust deployment. We propose \textbf{DiffE2E}, a diffusion-based end-to-end autonomous driving framework. This framework first performs multi-scale alignment of multi-sensor perception features through a hierarchical bidirectional cross-attention mechanism. It then introduces a novel class of hybrid diffusion-supervision decoders based on the Transformer architecture, and adopts a collaborative training paradigm that seamlessly integrates the strengths of both diffusion and supervised policy. DiffE2E models structured latent spaces, where diffusion captures the distribution of future trajectories and supervision enhances controllability and robustness. A global condition integration module enables deep fusion of perception features with high-level targets, significantly improving the quality of trajectory generation. Subsequently, a cross-attention mechanism facilitates efficient interaction between integrated features and hybrid latent variables, promoting the joint optimization of diffusion and supervision objectives for structured output generation, ultimately leading to more robust control. Experiments demonstrate that DiffE2E achieves state-of-the-art performance in both CARLA closed-loop evaluations and NAVSIM benchmarks. The proposed integrated diffusion-supervision policy offers a generalizable paradigm for hybrid action representation, with strong potential for extension to broader domains including embodied intelligence. More details and visualizations are available at \href{https://infinidrive.github.io/DiffE2E/}{project website}.
\end{abstract}

\section{Introduction}

End-to-end autonomous driving establishes a direct mapping from sensor data to control commands, effectively avoiding the error propagation issues in traditional modular architectures, significantly enhancing system decision-making efficiency and scenario adaptability \cite{hu2023planning,jiang2023vad,chen2024vadv2,sun2024sparsedrive,chen2024end,tampuu2020survey}. Current mainstream methods are based on explicit policy direct supervision framework\cite{shao2023safety,chitta2022transfuser,wu2022trajectory,jaeger2023hidden,zimmerlin2024hidden,shao2023reasonnet}, learning the mapping relationship between environmental observations and vehicle actions directly from massive driving data. However, they face two core challenges: the multimodal nature of driving behavior\cite{liao2024diffusiondrive} causes explicit supervised policy to easily produce suboptimal solutions, while the complexity of open scenarios leads to a sharp decline in generalization ability when facing data distribution shifts.

\begin{figure}
    \centering
    \includegraphics[width=\textwidth]{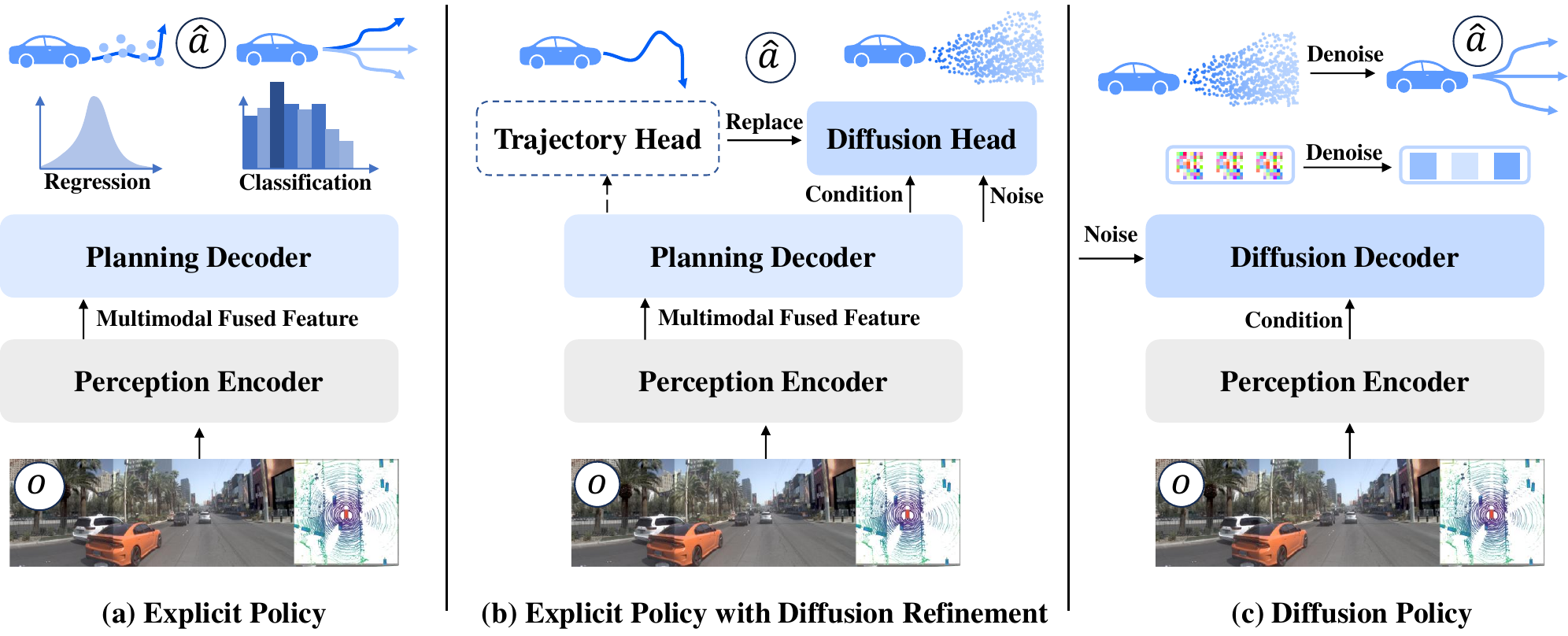}
    \vspace{-5pt}
    \caption{\small{\textbf{Comparison of end-to-end training paradigms.} (a) \textbf{Explicit Policy.} Directly predicts trajectories through supervised learning after sensor input processing. (b) \textbf{Explicit Policy with Diffusion Refinement.} Uses diffusion models to replace traditional explicit policy trajectory output heads. (c) \textbf{Diffusion Policy.} Uses diffusion models to directly generate trajectories based on perception encoder features.}}
    \label{fig:intro}
    \vspace{-5pt}
\end{figure}

The multimodal nature of driving leads to multiple reasonable decisions, but traditional supervised learning often averages them, resulting in suboptimal or unsafe behaviors \cite{dauner2024navsim}. Generalization is also a challenge, as end-to-end models can fail in unseen, complex scenarios. Together, these issues amplify safety risks, limiting the reliability of regression-based policy in real-world autonomous driving. Recent works \cite{chen2024vadv2,li2024hydra,li2025hydra,seff2023motionlm} tackle multimodality via discretized trajectory sets. However, such hard-coded methods break the continuity of decision-making and restrict adaptability, reducing generalization and downgrading continuous policy to fixed mode selection.

Diffusion models have demonstrated strong capabilities in modeling complex multimodal distributions and generating high-quality outputs, making them increasingly prominent in computer vision\cite{ho2020denoising,song2020denoising,song2020score,dhariwal2021diffusion}. This generation paradigm based on gradual denoising provides a new solution for addressing multimodal modeling and generalization problems in end-to-end autonomous driving. The successful application of diffusion models in robotic motion planning has already demonstrated their advantages in multimodal action sequence generation and long-term sequence prediction \cite{chi2023diffusion,ze20243d,janner2022planning}. However, applying diffusion models to autonomous driving systems faces unique challenges: autonomous driving needs to simultaneously meet multiple stringent requirements in open road environments, dealing with the uncertainty of highly dynamic traffic participants while ensuring real-time response; generating feasible trajectories that conform to road topology while ensuring traffic efficiency.

Recent works have explored diffusion models in autonomous driving planning \cite{zheng2025diffusion} and end-to-end control \cite{liao2024diffusiondrive,xing2025goalflow}, applying (Denoising Diffusion Implicit Model) DDIM \cite{song2020denoising}, DPM-Solver \cite{lu2022dpm}, and Rectified Flow \cite{liu2022flow} for trajectory generation, revealing the strong potential of generative approaches. However, most integrate diffusion models only after planning decoders, using them to replace explicit policy heads (see Figure \ref{fig:intro}(b)). This setup risks losing key perception features and constrains generation due to pre-processed decoder outputs. While some methods \cite{liao2024diffusiondrive} use trajectory anchors to enhance real-time performance, anchor-based designs can limit trajectory diversity. A more effective architecture is needed to fully harness the generative power of diffusion models.

To address these challenges, we propose \textbf{DiffE2E}, an innovative end-to-end autonomous driving framework, as shown in Figure \ref{fig:framework}. It first aligns LiDAR and image features via a hierarchical bidirectional cross-attention mechanism for accurate multi-scale perception. Based on this, we introduce a hybrid diffusion-supervision decoder based on the Transformer architecture and adopt a collaborative training mechanism that seamlessly integrates the advantages of diffusion policy and supervised policy. By combining diffusion models with explicit policy supervision, the latent space is structurally partitioned: on one hand, using diffusion models to model the distribution of future trajectories, effectively capturing their diversity and higher-order uncertainties; on the other hand, adopting explicit supervised learning strategies for fine-grained modeling of key control variables such as speed and surrounding vehicle motion information. A cross-attention mechanism enables interaction between integrated features and hybrid latent variables, supporting collaborative optimization and structured output between diffusion and explicit policy. Closed-loop tests in CARLA and evaluations in NAVSIM demonstrate that DiffE2E achieves state-of-the-art performance, ensuring safety, traffic efficiency, and strong generalization in complex scenarios.

In summary, our contributions are as follows:

1. We propose DiffE2E, the first end-to-end autonomous driving framework that uses diffusion models to directly generate trajectories and validate them in closed-loop testing in the CARLA simulator.

2. We introduce a hybrid diffusion-supervision decoder based on the Transformer architecture and adopt a collaborative training mechanism that seamlessly integrates the advantages of diffusion policy and supervised policy.

3. We conduct dual-platform benchmark testing, achieving state-of-the-art performance across multiple benchmarks in the CARLA simulator, and attaining a PDMS of 92.7 in the non-reactive simulation NAVSIM, while maintaining higher real-time performance compared to other methods.

\section{Preliminaries}
\textbf{Problem Definition:}  
This research focuses on end-to-end autonomous driving closed-loop control strategies based on diffusion models. The system directly takes multi-modal raw perception data as input, including front-view camera RGB images $\mathbf{I}_t \in \mathbb{R}^{H \times W \times 3}$, LiDAR point clouds $\mathbf{P}_t^\text{3D} \in \mathbb{R}^{N \times 3}$, and vehicle state information $\mathbf{s}_t \in \mathbb{R}^{d_s}$. The system outputs the ego vehicle's future trajectory $\bm{x}_0$, with its complete sampling distribution represented as:

\begin{equation}
    \bm{x}_0 = f_\theta(\mathbf{I}_t, \mathbf{P}_t, \mathbf{s}_t) = \int_{\bm{x}_{1:T}} p_\theta(\bm{x}_0|\bm{x}_1, \mathcal{C}) \prod_{t=2}^{T} p_\theta(\bm{x}_{t-1}|\bm{x}_t, \mathcal{C}) p(\bm{x}_T) \, d\bm{x}_{1:T}
\end{equation}

In diffusion modeling, $\bm{x}_t$ represents the intermediate variable at step $t$ in the diffusion process, with the final predicted trajectory being $\bm{x}_0 \in \mathbb{R}^{N \times d_c}$, where each path point $\mathbf{x}_i \in \mathbb{R}^{d_c}$ represents position information in the predicted trajectory. The condition information $\mathcal{C}$ is encoded from multi-modal sensor data through a cross-modal feature fusion module. Unlike traditional open-loop control, in closed-loop control, the trajectory decision at the current moment directly affects the perception input at the next moment, forming a dynamic feedback loop. This coupled relationship requires the model to have strong temporal consistency and robustness.

\textbf{Diffusion Models:}
Denoising Diffusion Probabilistic Models (DDPM) \cite{ho2020denoising} are powerful generative models that capture complex multimodal distributions via a two-phase process: forward diffusion gradually adds noise, while the reverse process reconstructs the data through iterative denoising. This framework naturally models the multimodality of driving behaviors. The forward process follows a Markov chain that transforms data $\bm{x}_0$ into noise over $T$ steps:
\begin{equation}
    q(\bm{x}_t|\bm{x}_{t-1}) = \mathcal{N}(\bm{x}_t; \sqrt{1-\beta_t}\bm{x}_{t-1}, \beta_t\bm{I})
\end{equation}

where $\{\beta_t\}_{t=1}^T$ controls noise levels. Through reparameterization, we can sample $\bm{x}_t$ directly:

\begin{equation}
    \bm{x}_t = \sqrt{\bar{\alpha}_t}\bm{x}_0 + \sqrt{1-\bar{\alpha}_t}\bm{\epsilon}, \quad \bm{\epsilon} \sim \mathcal{N}(0, \bm{I})
\end{equation}

where $\bar{\alpha}_t = \prod_{i=1}^{t}(1-\beta_i)$. While DDPM generates high-quality samples, its sequential process is computationally expensive. DDIM \cite{song2020denoising} address this with a non-Markovian process that accelerates generation while maintaining quality, making diffusion models more practical for real-time autonomous driving applications.

\section{Methodology}
\begin{figure}
    \centering
    \includegraphics[width=\textwidth]{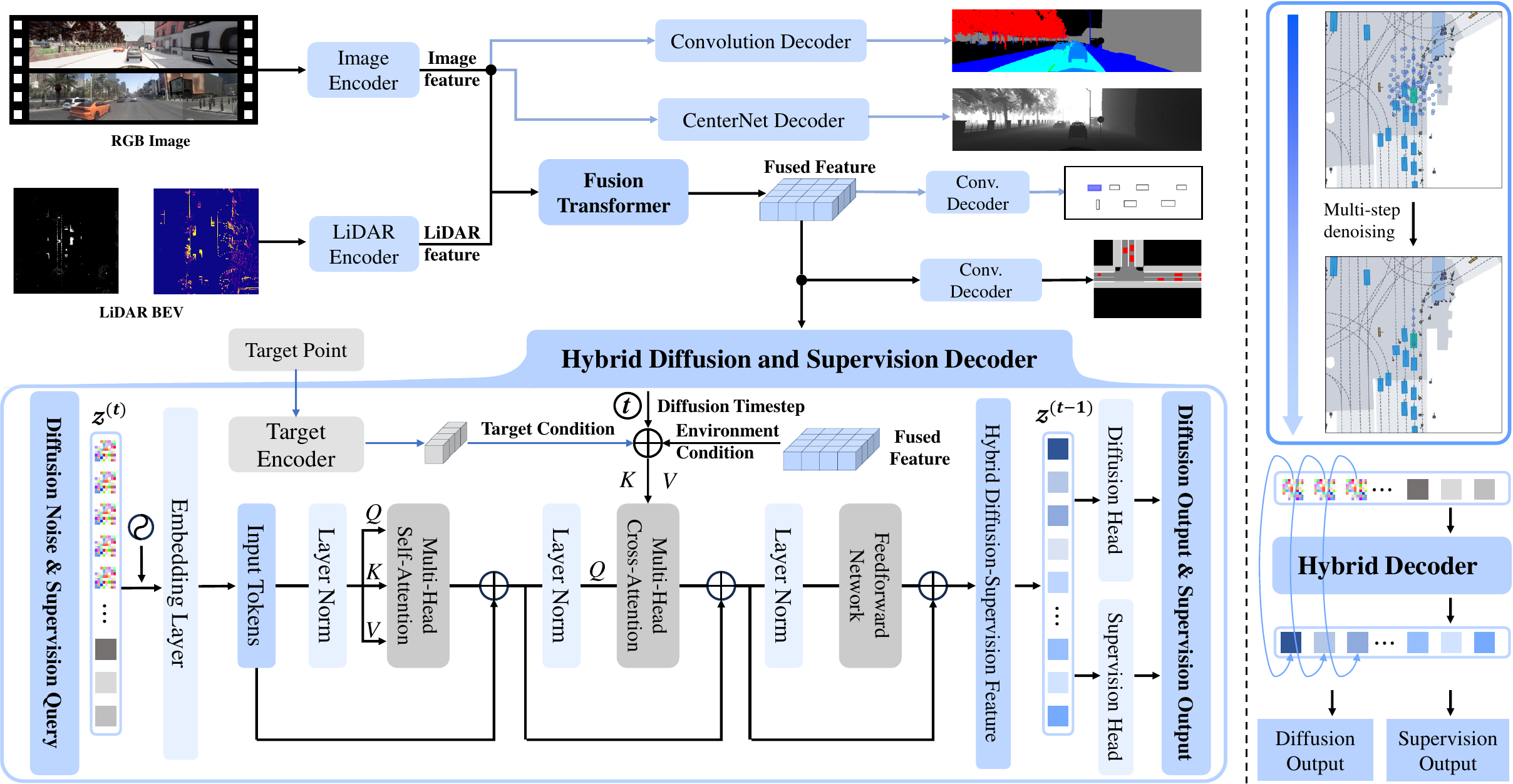}
    \vspace{-5pt}
    \caption{\small{\textbf{Overall architecture of DiffE2E.} The main architecture consists of a {Transformer-based perception module} and a {Hybrid Diffusion and Supervision Decoder}. The \textcolor{customblue}{blue arrows} ($\bm{\textcolor{customblue}{\rightarrow}}$) indicate the data flow exclusively used for the {CARLA} benchmark, while the \textbf{black arrows} ($\bm{\rightarrow}$) represent the data flow shared between both the {CARLA} and {NAVSIM} benchmarks.
    }}
    \label{fig:framework}
    \vspace{-5pt}
\end{figure}
\textbf{Overview:} DiffE2E is an end-to-end autonomous driving framework, illustrated in Figure~\ref{fig:framework}. It introduces a multimodal spatiotemporal fusion module during perception, aligning LiDAR and camera features via a hierarchical bidirectional cross-attention mechanism to build structured scene representations. In the decoding stage, a Transformer-based hybrid diffusion-supervision decoder is used with a collaborative training strategy that combines diffusion and supervised policy. A global condition integration module fuses scene features with target-related global context, and cross-attention enables effective interaction between these fused features and hybrid latent variables, allowing coordinated generation of diffusion trajectories and explicit outputs.

\subsection{Multimodal Fusion Perception Module}
The perception module aims to fuse multimodal sensor data to construct structured environmental representations. This paper adopts the Transfuser architecture as the basic perception backbone network\cite{jaeger2023hidden}, with inputs including wide-angle front-view RGB images $\mathbf{I}_t \in \mathbb{R}^{H \times W \times 3}$ and BEV representations $\mathbf{P}_t^\text{2D} \in \mathbb{R}^{H \times W \times C}$ constructed from raw LiDAR point clouds $\mathbf{P}_t^\text{3D} \in \mathbb{R}^{N \times 3}$. After extracting initial features from both branches, they enter a multi-scale cross-fusion module composed of multiple Transformer layers, achieving deep alignment and information interaction between LiDAR and image features through cross-modal attention mechanisms. Finally, the module outputs high-dimensional fusion features $\mathcal{C}$, global semantic representations, and image feature grids to support the fine-grained modeling requirements of downstream decision modules.

\subsection{Hybrid Diffusion and Supervision Module}
After the multimodal fusion perception module completes the integration of different sensor data, the proposed DiffE2E framework adopts an innovative architecture. By introducing a Transformer-based hybrid diffusion-supervision decoder and adopting a collaborative training mechanism, it seamlessly integrates the advantages of diffusion policy and supervised policy, as shown in Figure~\ref{fig:framework}. This section will provide a detailed explanation from different aspects including global condition integration, hybrid diffusion and supervision decoding, and decoder output module.

\textbf{Global Condition Integration:} To enhance the influence of target points in trajectory generation, they are used as global conditions \cite{jaeger2023hidden}. Target points are first projected via a linear layer $f_{\text{goal}}$ into a shared high-dimensional space, forming representations $\mathbf{g} \in \mathbb{R}^{\ell_g \times d}$. Meanwhile, diffusion timesteps $t$ are encoded into temporal embeddings $\mathbf{t}_{\text{emb}} \in \mathbb{R}^{\ell_t \times d}$ to help the model adapt across denoising stages. Finally, conditional features $\mathcal{C}$, goal features, and timestep embeddings are fused, and combined with learnable positional encoding ${\mathbf{E}_{\text{pos}}^{\mathcal{C}}}$ to form a contextual representation for trajectory decoding:
\begin{equation}
\tilde{\mathcal{C}} = \left[ \text{Concat}(\mathcal{C}, f_{\text{goal}}(\mathbf{g}), \mathbf{t}_{\text{emb}}) + \mathbf{E}_{\text{pos}}^{\mathcal{C}} \right] \in \mathbb{R}^{(\ell_c+\ell_g+\ell_t) \times d}
\end{equation}

This global condition integration mechanism incorporates target point information and timestep embeddings into perception features, enhancing the model's awareness of navigation goals and enabling dynamic adjustment of feature representations during denoising for more precise trajectory generation.

\textbf{Hybrid Diffusion and Supervision Decoding:}
If the trajectory length is $\ell_k$, this study initializes $\boldsymbol{\tau}_0 \in \mathbb{R}^{\ell_k \times d_c}$ to represent the ego vehicle's future trajectory. Using $d$ to represent the feature dimension of the embedding layer and $\ell_s$ to represent the feature length of the supervision task, the hybrid decoder designed in this study first maps the noisy trajectory $\boldsymbol{\tau}_t$ through a linear projection layer $f_{\text{enc}}$ to a high-dimensional feature representation, while concatenating the initialized query vectors $\mathbf{q}_0 \in \mathbb{R}^{\ell_s \times d}$ for supervision tasks, and adding learnable positional encoding $\mathbf{E}_{\text{pos}}^{\mathcal{Z}} \in \mathbb{R}^{(\ell_k+\ell_s) \times d}$ to obtain the initialized input vector:
\begin{equation}
    \mathcal{Z}_{\text{in}} = \left[ \text{Concat}(f_{\text{enc}}(\boldsymbol{\tau}_t), \mathbf{q}_0) + \mathbf{E}_{\text{pos}}^{\mathcal{Z}} \right] \in \mathbb{R}^{(\ell_k+\ell_s) \times d}, \quad t \in \mathbb{R}
\end{equation}

At each time step $t$ of the entire diffusion process, the input first passes through a multi-head self-attention layer to process the internal feature relationships of $\mathcal{Z}_{\text{in}}$:
\begin{equation}
    \mathcal{Z}_{\text{mid}} = \text{SelfAttn}(\mathcal{Z}_{\text{in}}), \quad \mathcal{Z}_{\text{mid}} \in \mathbb{R}^{(\ell_k+\ell_s) \times d}
\end{equation}

Then, through a cross-attention mechanism, $\mathcal{Z}_{\text{mid}}$ interacts with the conditional features $\tilde{\mathcal{C}}$, producing the final output:
\begin{equation}
    \mathcal{Z}_{\text{out}} = \text{CrossAttn}(\mathcal{Z}_{\text{mid}}, \tilde{\mathcal{C}}, \tilde{\mathcal{C}}), \quad
    \mathcal{Z}_{\text{out}} \in \mathbb{R}^{(\ell_k+\ell_s) \times d}, \ \tilde{\mathcal{C}} \in \mathbb{R}^{(\ell_c+\ell_g+\ell_t) \times d}
\end{equation}

where $\mathcal{Z}_\text{out}$ represents the feature vector after decoder output, $t$ is the current diffusion time step, $\tilde{\mathcal{C}}$ is the conditional feature integrated with target points, $\ell_c$ is the feature length obtained from the multimodal perception module, $\ell_g$ is the length of target point features, and $\ell_t$ is the length of diffusion time step features. Similarly, in the output feature $\mathcal{Z}_\text{out}$, the first $\ell_k$ positions correspond to the latent features for diffusion trajectory generation, while the last $\ell_s$ positions are latent features for supervision tasks. 

\textbf{Decoder Output Module:}
The decoder output module performs refined processing on the hybrid features $\mathcal{Z}_{\text{out}} \in \mathbb{R}^{(\ell_k+\ell_s) \times d}$, implementing a mixed decoding of diffusion generation and supervised learning. This module adopts a feature separation and task-specific decoding strategy, structurally decomposing the output features in the semantic space:
\begin{equation}
\begin{aligned}
\mathcal{Z}_{\text{diff}} &= \mathcal{Z}_{\text{out}}[:\ell_k] \in \mathbb{R}^{\ell_k \times d} \\
\mathcal{Z}_{\text{sup}} &= \mathcal{Z}_{\text{out}}[\ell_k:\ell_k+\ell_s] \in \mathbb{R}^{\ell_s \times d}
\end{aligned}
\end{equation}
where $\mathcal{Z}_{\text{diff}}$ encodes the high-dimensional latent representation of the diffusion trajectory, while $\mathcal{Z}_{\text{sup}}$ contains the structured feature information for supervised tasks.

\subsection{Diffusion and Supervision Collaborative Training Strategy}
Based on the hybrid diffusion and supervision decoder structure described above, this study proposes a collaborative training strategy based on diffusion generation and supervised learning. The core of this strategy lies in combining the generative capabilities of diffusion models with the precision of explicit supervision to form complementary advantages.

\textbf{Diffusion Loss Function:}  
\name{} employs a trajectory reconstruction-based loss function for diffusion generation, which directly optimizes the model's ability to recover the original trajectory from noisy inputs. Using $\mathcal{M}_\theta$ to represent the entire model, the loss function is formulated as:

\begin{equation}
\mathcal{L}_{\text{diff}} = \mathbb{E}_{t \sim \mathcal{U}(1,T),\, \mathbf{x}_0,\, \mathbf{x}_t \sim q_t(\mathbf{x}_t | \mathbf{x}_0)} \left[ \left\| \mathbf{x}_0 - \mathcal{M}_\theta(\mathbf{x}_t, t, \tilde{\mathcal{C}}) \right\|_2^2 \right]
\end{equation}

\textbf{Supervision Loss Function:}
The supervised learning loss adopts a multi-task combination optimization strategy, achieving refined gradient flow control and priority allocation through task-specific weight coefficients: $\mathcal{L}_{\text{sup}} = \sum_{i \in \Omega} \lambda_i \cdot \mathcal{L}_i(\mathbf{y}_i, \hat{\mathbf{y}}_i; \theta_i)$, where $\Omega$ represents the set of supervised tasks, $\lambda_i$ is the task weight, $\mathcal{L}_i(\mathbf{y}_i, \hat{\mathbf{y}}_i; \theta_i)$ represents the specific loss function for task $i$, $\mathbf{y}_i$ and $\hat{\mathbf{y}}_i$ are the ground truth labels and predicted values respectively, and $\theta_i$ represents the network parameters.

For example, for the speed prediction task in supervised learning, this study constructs a multi-category classification model based on semantic layering, including four speed states with clear physical meanings: \textit{braking}, \textit{walking speed}, \textit{slow}, and \textit{fast}. The classification prediction accuracy is optimized through a weighted cross-entropy loss function:

\begin{equation}
\mathcal{L}_{\text{speed}}(\mathbf{y}, \hat{\mathbf{p}}; \mathbf{w}) = -\frac{1}{N}\sum_{n=1}^{N}\sum_{i=1}^{4} w_i \cdot y_{n,i} \log(\hat{p}_{n,i} + \text{eps}), \quad \text{s.t.} \quad \sum_{i=1}^{4}\hat{p}_{n,i} = 1, \hat{p}_{n,i} \geq 0
\end{equation}

where $N$ is the batch size, $y_{n,i} \in {0,1}$ represents the ground truth label for the $n$-th sample in the $i$-th speed category (one-hot encoding), $\hat{p}_{n,i} \in [0,1]$ denotes the predicted normalized category probability, $w_i > 0$ is the balancing weight for category $i$ to mitigate data imbalance, and $\text{eps}$ is a numerical stability constant. Other supervised loss functions are detailed in Appendix \ref{app:loss_function}.

\section{Experiments}\label{sec:experiments}
\textbf{Experiment Setup:}
This research is primarily evaluated using the CARLA simulator closed-loop benchmark \cite{dosovitskiy2017carla} and the NAVSIM non-reactive simulation benchmark \cite{dauner2024navsim}. CARLA offers diverse urban scenes and sensor emulation, with its closed-loop mechanism providing real-time feedback to assess decision quality over long horizons. NAVSIM, built on OpenScene \cite{openscene2023} (a streamlined version of the nuPlan \cite{caesar2021nuplan} dataset),  provides 360° coverage via 8 cameras and 5 LiDARs, with 2Hz annotations of maps and object bounding boxes. Additional details are provided in Appendix \ref{app:experiment_setup}.

\subsection{Experiment on CARLA}
\textbf{Implementation Details:}
This study uses RegnetY-3.2GF \cite{radosavovic2020designing} as the encoder for image and LiDAR inputs. To optimize computation and training efficiency, a two-stage strategy is adopted: the first trains the perception module with multi-task losses; the second trains the diffusion decoder conditioned on the frozen perception outputs. We adopt CARLA Longest6, CARLA Town05 Long, and CARLA Town05 Short as evaluation benchmarks \cite{chitta2022transfuser,prakash2021multi}, using the official Driving Score (DS), Route Completion (RC), and Infraction Score (IS) as metrics. Detailed implementation details and baseline descriptions are provided in Appendix \ref{app:experiment_carla}.

\textbf{Main Results:}
As shown in Table \ref{tab:longest6}, our diffusion-based policy \name{} demonstrates excellent performance in the CARLA Longest6 benchmark. Among the three key evaluation metrics, \name{} ranks at the top: a DS of 83 (13.7\%higher than TF++WP), an IS of 0.86 (2.3\% above DriveAdapter+TCP), and a RC of 96, close to optimal. The slightly lower RC compared to TF++WP (96 vs. 97) reflects our emphasis on safety in high-risk scenarios. Further analysis shows that traditional explicit policy methods (e.g., LAV, Transfuser) struggle in complex scenarios, whereas the diffusion policy in \name{} better captures multimodal driving behaviors. Compared to TF++, which uses the same encoder (RegNetY-3.2GF) and input setup (C\&L), \name{} improves DS and IS by 20.3\% and 19.4\% respectively, with comparable RC (96 vs. 94). Overall, \name{} delivers robust, efficient end-to-end driving performance.

\begin{table}[t]
    \caption{\small \textbf{Comparison on the CARLA Longest6 benchmark.} “C \& L” denotes the use of both camera and LiDAR as sensor inputs. “Pri.” indicates the use of privileged information. The \textbf{best} and {\ul second best} results are highlighted in \textbf{bold} and {\ul underline}.}
    \label{tab:longest6}
    \resizebox{\textwidth}{!}{%
    \begin{tabular}{@{}l|ccc|
    >{\columncolor[HTML]{CDE8F8}}c cc@{}}
    \toprule
    \textbf{Method}         & \textbf{Traj. Decoder} & \textbf{Img. Encoder} & \textbf{Input} & \textbf{DS $\uparrow$} & \textbf{RC $\uparrow$} & \textbf{IS $\uparrow$} \\ \midrule
    WOR\cite{chen2021learning}               & -              & ResNet-18, 34\cite{he2016deep} & C    & 21 & 48          & 0.56 \\
    LAV v1\cite{chen2022learning}            & Explicit Policy & ResNet-18\cite{he2016deep}     & C\&L & 33 & 70          & 0.51 \\
    Interfuser\cite{shao2023safety}        & Explicit Policy & ResNet-50\cite{he2016deep}     & C\&L & 47 & 74          & 0.63 \\
    Transfuser\cite{chitta2022transfuser}        & Explicit Policy & RegNetY-3.2GF\cite{radosavovic2020designing} & C\&L & 47 & 93          & 0.50 \\
    TCP\cite{wu2022trajectory}               & Explicit Policy & ResNet-34\cite{he2016deep}     & C    & 48 & 72          & 0.65 \\
    LAV v2\cite{chen2022learning}            & Explicit Policy & ResNet-18, 34\cite{he2016deep} & C\&L & 58 & 83          & 0.68 \\
    Perception PlanT\cite{renz2022plant}  & Explicit Policy & -             & Pri. & 58 & 88          & 0.65 \\
    DriveAdapter\cite{jia2023driveadapter}      & Explicit Policy & ResNet-50\cite{he2016deep}     & C\&L & 59 & 82          & 0.68 \\
    ThinkTwice\cite{jia2023think}        & Explicit Policy & ResNet-50\cite{he2016deep}     & C\&L & 67 & 77          & 0.84 \\
    TF++\cite{jaeger2023hidden}              & Explicit Policy & RegNetY-3.2GF\cite{radosavovic2020designing} & C\&L & 69 & 94          & 0.72 \\
    DriveAdapter+TCP\cite{jia2023driveadapter} & Explicit Policy & ResNet-50\cite{he2016deep}     & C\&L & 71 & 88          & {\ul 0.85} \\
    TF++WP\cite{jaeger2023hidden}            & Explicit Policy & RegNetY-3.2GF\cite{radosavovic2020designing} & C\&L & {\ul 73} & \textbf{97} & 0.56 \\ \midrule
    \textbf{DiffE2E (Ours)} & Diffusion Policy               & RegNetY-3.2GF\cite{radosavovic2020designing}         & C\&L           & \textbf{83}            & {\ul 96}                     & \textbf{0.86}          \\ \bottomrule
    \end{tabular}%
    }
\end{table}

\begin{figure}
    \centering
    \includegraphics[width=\textwidth]{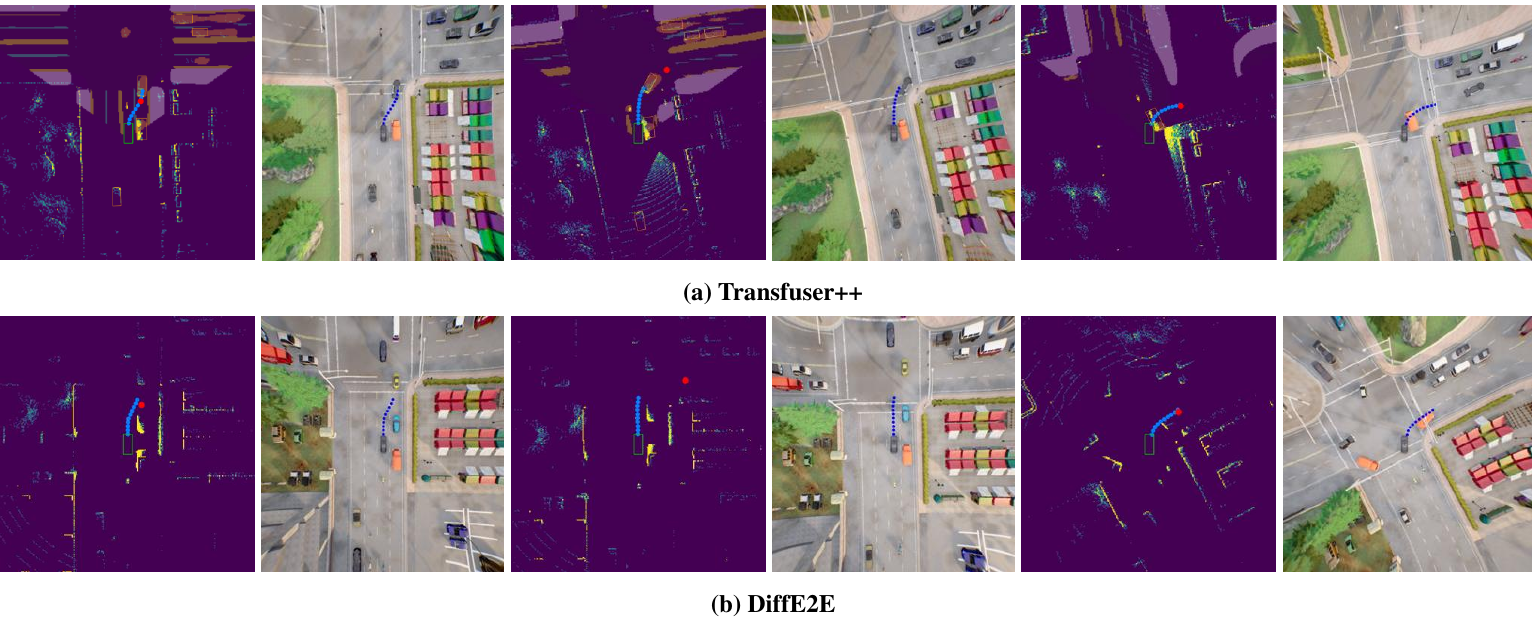}
    \vspace{-5pt}
    \caption{\small \textbf{Visualization in CARLA Simulator.} In both the LiDAR and scene visualizations, blue points represent the predicted trajectory, while red points in the LiDAR view denote the target waypoints.}
    \label{fig:carla}
    \vspace{-15pt}
\end{figure}

\textbf{Visualization:}
Figure \ref{fig:carla} shows a comparison in a typical right-turn scenario. Initially, both TF++ and \name{} plan similar paths by first merging right. When a vehicle appears, TF++ sticks to its preset path and collides, while \name{} adapts by temporarily going forward, then safely merging after the vehicle passes. This demonstrates \name{}'s superior multimodal generation capability and and real-time adaptability in dynamic traffic, effectively avoiding collisions.

\subsection{Experiment on NAVSIM}

\textbf{Implementation Details:}
This study builds a model training framework based on NAVSIM's navtrain dataset. Unlike the CARLA setup, we adopt VovNetV2-99\cite{lee2019energy} as the feature extraction backbone network in NAVSIM. The Predictive Driver Model Score (PDMS) is used as a comprehensive metric, combining key driving dimensions via weighted integration: No at-fault Collision (NC), Drivable Area Compliance (DAC), Time-To-Collision (TTC), Comfort (C), and Ego Progress (EP). Detailed implementation details and baseline descriptions can be found in Appendix \ref{app:experiment_navsim}.

\textbf{Main Results:}
As shown in Table \ref{tab:navsim}, \name{} achieves excellent overall performance on the NAVSIM benchmark, with a PDMS score of 92.7—outperforming Hydra-MDP++ (91.0), GoalFlow (90.3), and DiffusionDrive (88.1). This highlights the strength of our diffusion-based end-to-end approach in multi-dimensional driving evaluation. In safety and compliance, \name{} excels: it achieves a no-fault collision rate of 99.9 (vs. 98.6 for Hydra-MDP++ and 98.4 for GoalFlow), and shares the top drivable area compliance score of 98.6 with Hydra-MDP++. On time-to-collision, \name{} leads significantly at 99.3, 4.2 points above Hydra-MDP++. For efficiency and comfort, \name{} scores 85.3 in ego progress (second only to Hydra-MDP++'s 85.7), and achieves 99.9 in driving comfort—close to the best-performing methods—indicating smooth, human-like trajectories.

Additionally, \name{} adopts a complete diffusion decoder paradigm, rather than the diffusion replanning method after the Transfuser planning decoder used by DiffusionDrive and GoalFlow, indicating our method has stronger expressive capability in modeling policy space. Furthermore, \name{} achieves leading or near-optimal results on all key NAVSIM metrics, demonstrating its robustness, safety, and practicality and offering a new paradigm for end-to-end autonomous driving.

\begin{figure}
    \centering
    \includegraphics[width=\textwidth]{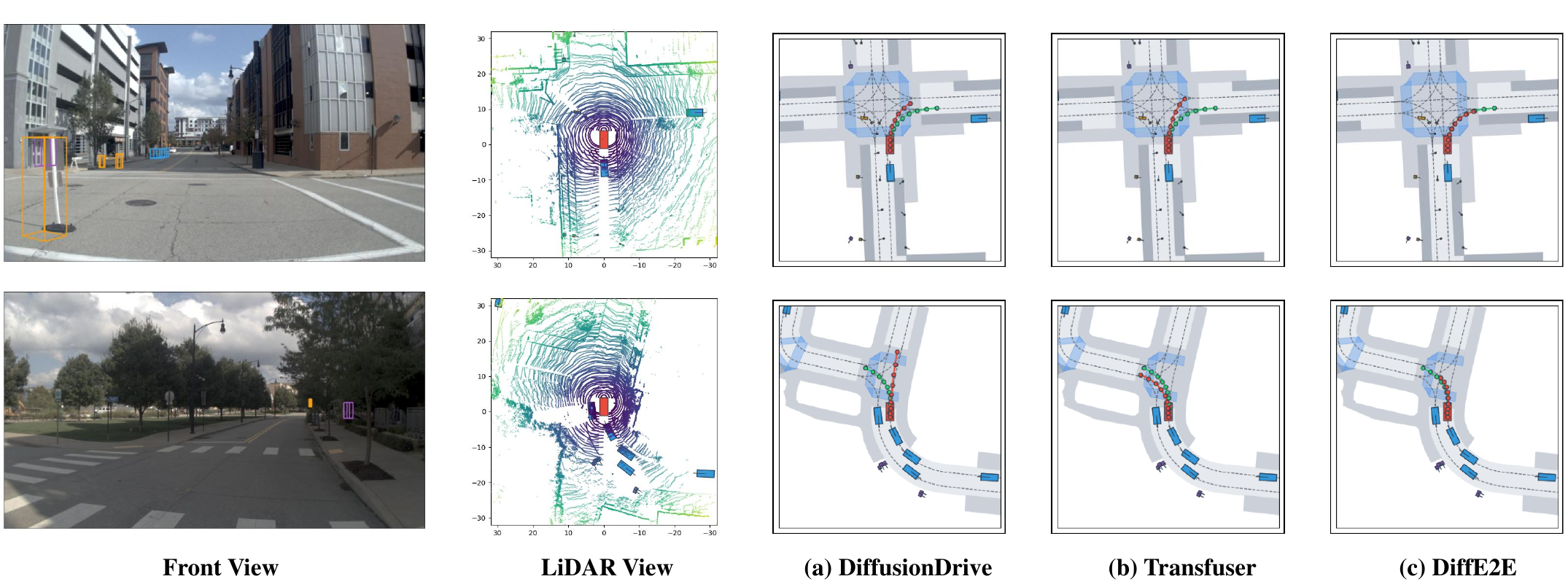}
    \vspace{-5pt}
    \caption{\small \textbf{Visualization in Navtest benchmark.} {Red trajectories} denote the predicted paths of each method, while {green trajectory} corresponds to the ground truth.}
    \label{fig:navsim}
    \vspace{-15pt}
\end{figure}

\textbf{Visualization:}
To validate the generalization ability and superiority of \name{}, we selected two representative complex driving scenarios for comparative analysis (Figure \ref{fig:navsim}). Green trajectories denote human references, and red ones indicate planned trajectories. In right-turn intersections, baseline methods often deviate or cross boundaries, while \name{} accurately follows lane edges with smooth turns. In small intersection left-turns, DiffusionDrive misinterprets navigation intent and plans a straight trajectory, Transfuser incorrectly chooses the right lane, while only \name{} accurately executes the left-turn instruction with a trajectory almost completely matching the reference. This demonstrates \name{}'s accuracy and safety in trajectory planning. More visualization results can be found in \Cref{app:visualization}.

\begin{table}[t]
    \caption{\small \textbf{Comparison on the Navtest benchmark.} “S” denotes using only the ego vehicle state as input. * indicates that these two methods use diffusion models as a post-planning decoder, while \name{} employs a diffusion model as the entire decoder to generate trajectories.}
    \label{tab:navsim}
    \resizebox{\textwidth}{!}{%
    \begin{tabular}{@{}lccc
    >{\columncolor[HTML]{CDE8F8}\centering\arraybackslash}c
    ccccc@{}}
    \toprule
    \multicolumn{1}{l|}{\textbf{Method}} &
      \textbf{Traj. Decoder} &
      \textbf{Img. Enc.} &
      \multicolumn{1}{c|}{\textbf{Input}} &
      \textbf{PDMS$\uparrow$} &
      \textbf{NC$\uparrow$} &
      \textbf{DAC$\uparrow$} &
      \textbf{EP$\uparrow$} &
      \textbf{TTC$\uparrow$} &
      \textbf{C$\uparrow$} \\ \midrule
    \multicolumn{1}{l|}{Human}          & -              & -         & \multicolumn{1}{c|}{-}    & 94.8 & 100  & 100  & 87.5 & 100  & 99.9 \\ \midrule
    \multicolumn{1}{l|}{AD-MLP\cite{zhai2023rethinking}}         & Explicit Policy & -         & \multicolumn{1}{c|}{S}    & 65.6 & 93.0 & 77.3 & 62.8 & 83.6 & \textbf{100} \\
    \multicolumn{1}{l|}{VADv2\cite{chen2024vadv2}}          & Explicit Policy & ResNet-34\cite{he2016deep} & \multicolumn{1}{c|}{C\&L} & 80.9 & 97.2 & 89.1 & 76.0 & 91.6 & \textbf{100} \\
    \multicolumn{1}{l|}{UniAD\cite{hu2023planning}}          & Explicit Policy & ResNet-34\cite{he2016deep} & \multicolumn{1}{c|}{C}    & 83.4 & 97.8 & 91.9 & 78.8 & 92.9 & \textbf{100} \\
    \multicolumn{1}{l|}{LTF\cite{chitta2022transfuser}}            & Explicit Policy & ResNet-34\cite{he2016deep} & \multicolumn{1}{c|}{C}    & 83.8 & 97.4 & 92.8 & 79   & 92.4 & \textbf{100} \\
    \multicolumn{1}{l|}{PARA-Drive\cite{weng2024drive}}     & Explicit Policy & ResNet-34\cite{he2016deep} & \multicolumn{1}{c|}{C}    & 84.0 & 97.9 & 92.4 & 79.3 & 93   & 99.8 \\
    \multicolumn{1}{l|}{Transfuser\cite{chitta2022transfuser}}     & Explicit Policy & ResNet-34\cite{he2016deep} & \multicolumn{1}{c|}{C\&L} & 84.0 & 97.7 & 92.8 & 79.2 & 92.8 & \textbf{100} \\
    \multicolumn{1}{l|}{LAW\cite{li2024enhancing}}            & Explicit Policy & ResNet-34\cite{he2016deep} & \multicolumn{1}{c|}{C}    & 84.6 & 96.4 & 95.4 & 81.7 & 88.7 & {\ul 99.9} \\
    \multicolumn{1}{l|}{DRAMA\cite{yuan2024drama}}          & Explicit Policy & ResNet-34\cite{he2016deep} & \multicolumn{1}{c|}{C\&L} & 85.5 & 98   & 93.1 & 80.1 & 94.8 & \textbf{100} \\
    \multicolumn{1}{l|}{Hydra-MDP\cite{li2024hydra}}      & Explicit Policy & ResNet-34\cite{he2016deep} & \multicolumn{1}{c|}{C\&L} & 86.5 & 98.3 & 96.0 & 78.7 & 94.6 & \textbf{100} \\
    \multicolumn{1}{l|}{DiffusionDrive\cite{liao2024diffusiondrive}} & Diffusion Policy*     & ResNet-34\cite{he2016deep} & \multicolumn{1}{c|}{C\&L} & 88.1 & 98.2 & 96.2 & 82.2 & 94.7 & \textbf{100} \\
    \multicolumn{1}{l|}{GoalFlow\cite{xing2025goalflow}}       & Diffusion Policy*     & V2-99\cite{lee2019energy}     & \multicolumn{1}{c|}{C\&L} & 90.3 & 98.4 & {\ul 98.3} & 85   & 94.6 & \textbf{100} \\
    \multicolumn{1}{l|}{Hydra-MDP++\cite{li2025hydra}} &
      Explicit Policy &
      V2-99\cite{lee2019energy} &
      \multicolumn{1}{c|}{C} &
      {\ul 91.0} &
      {\ul 98.6} &
      \textbf{98.6} &
      \textbf{85.7} &
      {\ul 95.1} &
      \textbf{100} \\ \midrule
    \multicolumn{1}{l|}{\textbf{DiffE2E (Ours)}} &
      Diffusion Policy &
      V2-99\cite{lee2019energy} &
      \multicolumn{1}{c|}{C\&L} &
      \textbf{92.7} &
      \textbf{99.9} &
      \textbf{98.6} &
      {\ul 85.3} &
      \textbf{99.3} &
      {\ul 99.9} \\ \bottomrule
    \end{tabular}%
    }
\end{table}

\subsection{Ablation Studies}
\begin{figure*}[t]
    \vspace{-5pt}
    \begin{minipage}[h]{.45\textwidth}
        \centering
        \vspace{-10pt}
        \captionof{table}{\small Ablation results of DiffE2E on CARLA Longest6 benchmark.}
        \label{tab:ablation}
        \scriptsize
        \setlength{\tabcolsep}{4pt}
        \begin{tabular}{@{}ll
            >{\columncolor[HTML]{CDE8F8}}l ll@{}}
            \toprule
            \textbf{Type}           & \textbf{Method}     & \textbf{DS}   & \textbf{RC}   & \textbf{IS}   \\ \midrule
            Base                    & \textbf{DiffE2E}    & \textbf{82.9} & \textbf{96.2} & \textbf{0.86} \\ \midrule
                                    & w/o ego state       & 68.9          & 88.8          & 0.81          \\
            \multirow{-2}{*}{Input} & w/o command         & 69.6          & 93.8          & 0.77          \\ \midrule
            Component               & w/o GRU             & 66.8          & 75.5          & 0.88          \\ \midrule
                                    & Full Diffusion      & 70.1          & 91.1          & 0.76          \\
                                    & Full Discrimination & 70.3          & 83.5          & 0.86          \\
                                    & One-stage Training  & 18.2          & 21.8          & 0.79          \\
            \multirow{-4}{*}{Training Paradigm} & \textbf{Two-stage Training} & \textbf{82.9} & \textbf{96.2} & \textbf{0.86} \\ \midrule
                                    & Noise Prediction               & 20.1          & 27.2          & 0.72          \\
            \multirow{-2}{*}{Output}            & \textbf{Trajectory Reconstruction}         & \textbf{82.9} & \textbf{96.2} & \textbf{0.86} \\ \bottomrule
            \end{tabular}
    \end{minipage}
    \hfill
    \begin{minipage}[h]{.45\textwidth}
    \centering
    \vspace{10pt}
    \includegraphics[width=0.98\textwidth]{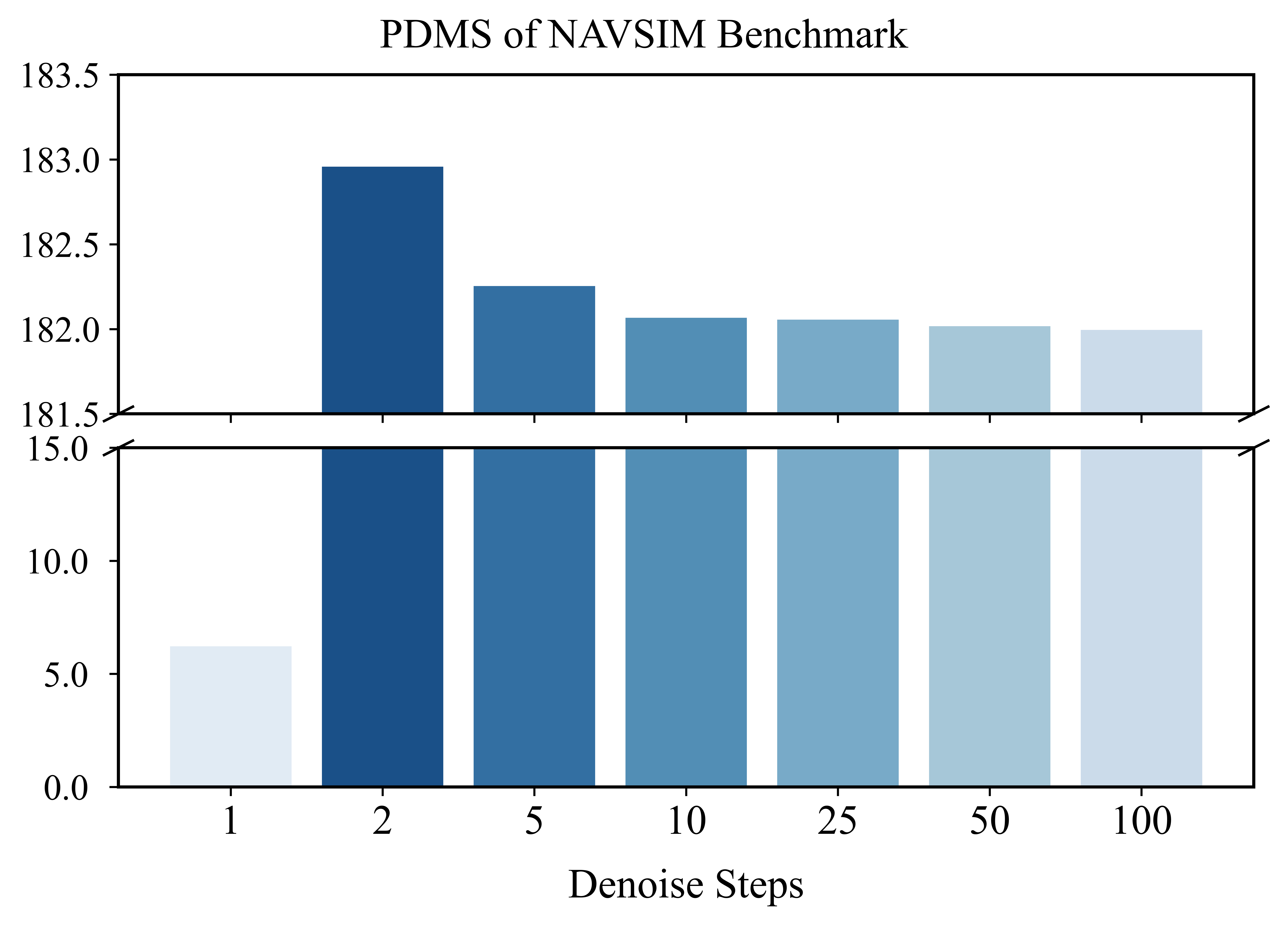}
    \caption{\small Ablation study of denoising steps on Navtest benchmark.}
    \label{fig:diff_steps}
    \end{minipage}
\end{figure*}

To evaluate the contribution of each component in the DiffE2E framework, we conducted a series of ablation experiments (Table \ref{tab:ablation}). Although NAVSIM testing is faster, we used the CARLA environment for its more reliable closed-loop evaluations. For model inputs, we ablated the ego state and navigation instructions separately. In both cases, driving scores dropped, confirming the importance of ego state for accurate planning and navigation input for intention understanding. Regarding architecture, removing the GRU module led to a significant score drop, showing its role in improving prediction in complex scenarios. In terms of training, we compared hybrid diffusion, full diffusion, and explicit policy paradigms, along with one-stage vs. two-stage strategies. Both full diffusion and explicit policy training reduced performance, validating the hybrid approach. One-stage training achieved only a driving score of 18.2—78\% lower than two-stage—resulting in poor lane-keeping. This indicates that joint training of perception and planning poses challenges, while two-stage training enables each module to be optimized effectively. Further discussion on output types is provided in Appendix \ref{app:RQ1}.

Additionally, we conducted ablation studies on the number of denoising steps in the diffusion model (Figure \ref{fig:diff_steps}). Due to CARLA's randomness and the minor effect of denoising steps, we used the more stable NAVSIM Navtest Benchmark. For clarity, we set 92.705 as the zero baseline and applied a 1e4 scaling factor. Results show PDMS is lowest at 1 step (incomplete denoising), peaks at 2 steps, then gradually declines, indicating full denoising at that point. Thus, 2 denoising steps are used in DiffE2E to balance performance and real-time efficiency, which is essential for autonomous driving. The discussion on real-time performance can be found in \Cref{app:RQ2}.

\vspace{-10pt}

\section{Related Works}
\textbf{End-to-end Autonomous Driving:} End-to-end autonomous driving has advanced significantly in multimodal perception fusion and decision-making.  UniAD\cite{hu2023planning} constructs a full-stack Transformer to coordinate perception-prediction-planning tasks, VAD\cite{jiang2023vad} designs vectorized scene representation to improve planning efficiency, VADv2\cite{chen2024vadv2} models action space distribution through a trajectory vocabulary library, SparseDrive\cite{sun2024sparsedrive} proposes sparse trajectory representation for efficient driving without BEV, and Hydra-MDP series\cite{li2024hydra,li2025hydra} designs a multi-teacher distillation framework to integrate rule-based systems with human driving knowledge. Transfuser\cite{chitta2022transfuser} fuses camera and LiDAR features via Transformer for intersection decisions; TCP\cite{wu2022trajectory} jointly trains trajectory and control predictions; InterFuser\cite{shao2023safety} introduces safety thinking maps for multi-view multimodal fusion, and TF++\cite{jaeger2023hidden} enhances the decoder and proposes decoupled speed prediction. However, these explicitly supervised methods often reduce multimodal driving behaviors to a single deterministic output, resulting in averaged and suboptimal decisions in multi-choice scenarios. While effective in specific cases, they generalize poorly to complex conditions beyond the training data, limiting their ability to handle the diversity of the open world.

\textbf{Diffusion Model in Transportation and Autonomous Driving:}
Diffusion models are profoundly transforming the transportation and autonomous driving fields with their excellent multimodal generation capabilities. In transportation, Diffusion-ES\cite{yang2024diffusion} innovatively combines evolutionary strategies with diffusion models, achieving complex driving behavior generation without requiring differentiable reward functions, with its zero-shot performance significantly surpassing traditional methods in the nuPlan benchmark. VBD\cite{huang2024versatile} uses game theory to guide adversarial scenario generation, enhancing simulation realism. MotionDiffuser\cite{jiang2023motiondiffuser} introduces a permutation-invariant architecture for constrained multi-agent trajectory sampling, ensuring interaction consistency. Diffusion Planner\cite{zheng2025diffusion} leverages DPM-Solver\cite{lu2022dpm} and classifier guidance for fast, safe, and personalized trajectory generation in closed-loop planning. However, most of these methods are based on perfect perception assumptions, ignoring the impact of state estimation errors caused by perception uncertainties in practical applications. In the field of end-to-end autonomous driving, although the application of diffusion models has achieved preliminary results, it still faces many challenges. DiffusionDrive\cite{liao2024diffusiondrive} first introduced diffusion for end-to-end driving with an anchored strategy. HE-Drive\cite{wang2024he} uses conditional DDPM\cite{ho2020denoising} and vision-language models for scoring, producing human-like, spatiotemporally consistent trajectories at high computational cost. GoalFlow\cite{xing2025goalflow} addresses trajectory divergence via goal-driven flow matching and efficient one-step generation. These works demonstrate the enormous potential of diffusion models in the field of end-to-end autonomous driving.

\section{Conclusion}\label{sec:conclusion}
This research proposes an innovative end-to-end autonomous driving framework named DiffE2E, which integrates a Transformer-based hybrid diffusion-supervision decoder and introduces a collaborative training mechanism, effectively combining the advantages of diffusion and supervised policy. We designed a structured latent space modeling method: using diffusion models to model future trajectory distributions, capturing the diversity and uncertainty of behaviors; while introducing explicit supervision for fine-grained modeling of key control variables such as speed and surrounding vehicle dynamics, enhancing perception capabilities for physical constraints and environmental changes, thereby improving the controllability and precision of predictions. In both CARLA closed-loop testing and NAVSIM non-reactive simulation, DiffE2E achieved leading performance, balancing traffic efficiency and safety, while demonstrating excellent generalization capabilities.

\bibliography{neurips_2025}
\bibliographystyle{plainnat}


\newpage
\appendix

\section{Details of Multi-task Loss Function Design}\label{app:loss_function}
\subsection{Loss Function Design}
Compared to the deterministic mapping supervision learning paradigm of explicit policy, diffusion models need to perform denoising tasks at multiple noise levels, presenting higher optimization complexity and computational overhead. To improve training efficiency and ensure model convergence stability in the CARLA simulator environment, this research constructs a hierarchical two-stage training strategy: the first stage focuses on optimizing parameters of the multimodal perception module, while the second stage adopts an end-to-end joint training paradigm, simultaneously optimizing parameter distributions of both the perception module and diffusion decoder.

The first stage employs a multi-task learning framework, designing specific supervision signals for core perception subtasks (including high-precision semantic segmentation, monocular depth estimation, and 3D object detection\cite{jaeger2023hidden}), and constructing a composite loss function through an adaptive weighting mechanism to achieve efficient parameter learning for the perception module. This research further designs a modular loss function system, with each function component defined as follows:

\begin{itemize}[leftmargin=*]
    \item \textbf{Image Semantic Segmentation Loss} $\mathcal{L}_{\text{sem}}$: Supervises semantic segmentation predictions from the image perspective, using a class-weighted cross-entropy loss function:
    \begin{equation}
        \mathcal{L}_{\text{sem}}(\mathbf{y}, \hat{\mathbf{y}}; \mathbf{w}) = -\sum_{c=1}^{C} w_c \cdot \sum_{i=1}^{H \times W} y_{i,c} \log(\hat{y}_{i,c} + \text{eps})
    \end{equation}
    where $C$ is the total number of semantic classes, $H \times W$ is the image resolution, $y_{i,c} \in \{0,1\}$ indicates whether pixel $i$ belongs to class $c$ in the ground truth, $\hat{y}_{i,c} \in [0,1]$ is the predicted probability, $w_c > 0$ is the balancing weight for class $c$, and $\text{eps}$ is a constant for numerical stability.

    \item \textbf{Bird's-Eye View Semantic Segmentation Loss} $\mathcal{L}_{\text{bev}}$: For semantic predictions in BEV space, cross-entropy loss is calculated only within the camera-visible region $\mathcal{M}_{\text{valid}}$:
    \begin{equation}
        \mathcal{L}_{\text{bev}}(\mathbf{S}_{\text{bev}}, \hat{\mathbf{S}}_{\text{bev}}; \mathcal{M}_{\text{valid}}) = -\frac{1}{|\mathcal{M}_{\text{valid}}|}\sum_{(x,y) \in \mathcal{M}_{\text{valid}}}\sum_{c=1}^{C} S_{\text{bev}}(x,y,c) \log(\hat{S}_{\text{bev}}(x,y,c) + \text{eps})
    \end{equation}
    where $\mathcal{M}_{\text{valid}}$ represents the set of valid pixels, $|\mathcal{M}_{\text{valid}}|$ is the number of valid pixels, $S_{\text{bev}}(x,y,c)$ and $\hat{S}_{\text{bev}}(x,y,c)$ are the ground truth label and predicted probability for class $c$ at BEV coordinate $(x,y)$, respectively.

    \item \textbf{Depth Estimation Loss} $\mathcal{L}_{\text{depth}}$: As an auxiliary task for the image branch, masked $\ell_1$ loss is used for depth supervision:
    \begin{equation}
        \mathcal{L}_{\text{depth}}(\mathbf{D}, \hat{\mathbf{D}}; \mathcal{M}_{\text{depth}}) = \frac{1}{|\mathcal{M}_{\text{depth}}|}\sum_{(u,v) \in \mathcal{M}_{\text{depth}}} |D(u,v) - \hat{D}(u,v)|
    \end{equation}
    where $\mathcal{M}_{\text{depth}}$ is the set of valid depth pixels, $D(u,v)$ and $\hat{D}(u,v)$ are the ground truth depth value and predicted depth value at pixel $(u,v)$, respectively.

    \item \textbf{Object Detection Loss} $\mathcal{L}_{\text{det}}$: This loss function combines multiple subtasks to jointly optimize key attributes of objects, including position, size, orientation, velocity, and braking status, to enhance overall detection accuracy and temporal consistency. It includes weighted combinations of several subtask loss functions. Each subloss function is defined as follows:
    
    \begin{itemize}[label=\textasteriskcentered]
        \item \textbf{Center Heatmap Loss} $\mathcal{L}_{\text{hm}}$: Uses Gaussian focal loss to optimize object center point prediction:
        \begin{equation}
            \mathcal{L}_{\text{hm}}(\mathbf{P}, \hat{\mathbf{P}}; \alpha, \gamma) = \frac{1}{N}\sum_{(x,y) \in \Omega} 
            \begin{cases}
                (1-\hat{P}_{x,y})^{\alpha}\log(\hat{P}_{x,y}), & \text{if } P_{x,y}=1 \\
                (1-P_{x,y})^{\gamma}(\hat{P}_{x,y})^{\alpha}\log(1-\hat{P}_{x,y}), & \text{otherwise}
            \end{cases}
        \end{equation}
        where $N$ is the number of valid objects, $\Omega$ is the feature map space, $P_{x,y}$ and $\hat{P}_{x,y}$ are the ground truth heatmap value and predicted heatmap value at position $(x,y)$, respectively, and $\alpha$ and $\gamma$ are focal loss modulation parameters.
        
        \item \textbf{Object Size Loss} $\mathcal{L}_{\text{wh}}$: Uses $\ell_1$ loss to optimize object width and height prediction:
        \begin{equation}
            \mathcal{L}_{\text{wh}}(\mathbf{W}, \hat{\mathbf{W}}) = \frac{1}{N}\sum_{i=1}^{N} \|\mathbf{W}_i - \hat{\mathbf{W}}_i\|_1
        \end{equation}
        where $\mathbf{W}_i = [w_i, h_i]^T$ and $\hat{\mathbf{W}}_i = [\hat{w}_i, \hat{h}_i]^T$ are the ground truth size and predicted size of the $i$-th object, respectively.
        
        \item \textbf{Center Point Offset Loss} $\mathcal{L}_{\text{off}}$: Also uses $\ell_1$ loss to optimize fine-grained center point position:
        \begin{equation}
            \mathcal{L}_{\text{off}}(\mathbf{O}, \hat{\mathbf{O}}) = \frac{1}{N}\sum_{i=1}^{N} \|\mathbf{O}_i - \hat{\mathbf{O}}_i\|_1
        \end{equation}
        where $\mathbf{O}_i = [o_{x,i}, o_{y,i}]^T$ and $\hat{\mathbf{O}}_i = [\hat{o}_{x,i}, \hat{o}_{y,i}]^T$ are the ground truth offset and predicted offset of the $i$-th object, respectively.
        
        \item \textbf{Yaw Classification Loss} $\mathcal{L}_{\text{cls}}$: Uses cross-entropy loss to optimize yaw angle category prediction:
        \begin{equation}
            \mathcal{L}_{\text{cls}}(\mathbf{Y}, \hat{\mathbf{Y}}) = \frac{1}{N}\sum_{i=1}^{N} -\log\left(\frac{\exp(\hat{Y}_{i,c_i})}{\sum_{j=1}^{K}\exp(\hat{Y}_{i,j})}\right)
        \end{equation}
        where $c_i$ is the ground truth yaw angle category of the $i$-th object, $K$ is the total number of yaw angle categories, and $\hat{Y}_{i,j}$ is the predicted score of the $i$-th object belonging to the $j$-th category.
        
        \item \textbf{Yaw Residual Loss} $\mathcal{L}_{\text{res}}$: Uses smooth $\ell_1$ loss to optimize yaw angle residual prediction:
        \begin{equation}
            \mathcal{L}_{\text{res}}(\mathbf{R}, \hat{\mathbf{R}}; \delta) = \frac{1}{N}\sum_{i=1}^{N} \text{smooth}_{L1}(R_i - \hat{R}_i; \delta)
        \end{equation}
        where $R_i$ and $\hat{R}_i$ are the ground truth yaw angle residual and predicted yaw angle residual of the $i$-th object, respectively, and $\text{smooth}_{L1}$ is the smooth $\ell_1$ loss function. The smooth L1 loss function is defined as:

        \begin{equation}
        \text{smooth}_{L1}(z; \delta) = 
        \begin{cases}
        0.5 \cdot z^2 / \delta, & \text{if } |z| < \delta \\
        |z| - 0.5 \cdot \delta, & \text{otherwise}
        \end{cases}
        \end{equation}
    \end{itemize}
\end{itemize}

The above loss functions form the core components of the first-stage training for the perception module. By adopting a multi-task learning paradigm, the model can simultaneously reconstruct multiple scene feature representations, including semantic segmentation maps, depth maps, and bird's-eye view segmentation. This approach not only achieves efficient and comprehensive scene understanding but also lays a solid perceptual foundation for subsequent trajectory generation tasks, ensuring that downstream decision-making modules can perform precise reasoning based on high-quality scene representations.

Additionally, on the NAVSIM benchmark, we use agent detection as an auxiliary task to enhance the model's perception capabilities in complex driving scenarios. The agent detection task adopts a hierarchical loss structure, including two complementary subtasks: existence discrimination and geometric parameter regression:

\begin{itemize}[leftmargin=*]
\item \textbf{Agent Existence Classification Loss:} Through binary cross-entropy loss, the model's ability to determine the presence of agents in the scene is optimized:
\begin{equation}
\mathcal{L}_{\text{exist}}(\mathbf{y}, \hat{\mathbf{p}}; \alpha, \beta) = -\frac{1}{N}\sum_{n=1}^{N} \left[ \alpha \cdot y_{n} \cdot (1-\hat{p}_{n})^{\beta} \log(\hat{p}_{n}) + (1-\alpha) \cdot (1-y_{n}) \cdot \hat{p}_{n}^{\beta} \log(1-\hat{p}_{n}) \right]
\end{equation}

where $y_{n} \in \{0,1\}$ represents the ground truth label of whether an agent exists in the $n$-th sample, $\hat{p}_{n} \in [0,1]$ is the predicted existence probability, and $\alpha \in [0,1]$ and $\beta \geq 0$ are focal loss modulation parameters used to adjust the gradient contribution ratio of easy and difficult samples, increasing the model's attention to difficult samples.

\item \textbf{Agent Geometry Parameter Regression Loss:} For existing agents, an adaptively weighted smooth L1 loss function is used to accurately regress their spatial position and morphological parameters:
\begin{equation}
\mathcal{L}_{\text{box}}(\mathbf{b}, \hat{\mathbf{b}}; \boldsymbol{\mu}, \delta) = \frac{1}{M}\sum_{m=1}^{M} \sum_{i \in \{x,y,w,h,\theta\}} \mu_i \cdot \text{smooth}_{L1}(b_{m,i} - \hat{b}_{m,i}; \delta)
\end{equation}

where $M$ is the number of samples with agents in the batch, $b_{m,i}$ and $\hat{b}_{m,i}$ represent the ground truth value and predicted value of the $i$-th parameter of the agent bounding box in the $m$-th sample, respectively, $\boldsymbol{\mu} = [\mu_x, \mu_y, \mu_w, \mu_h, \mu_{\theta}]^T$ is the importance weight vector for each parameter, and $\delta$ is the threshold parameter for the smooth L1 loss. This hierarchical loss design can simultaneously optimize the classification accuracy and localization precision of agent detection, ensuring accurate regression of bounding box parameters while maintaining high recall rates.
\end{itemize}

\begin{table}[]
    \caption{Hyperparameters of \name{} on CARLA.}
    \label{tab:hyper_carla}
    \resizebox{\textwidth}{!}{%
    \begin{tabular}{@{}l|l|lc@{}}
    \toprule
    \textbf{Benchmark} & \textbf{Type}              & \textbf{Parameter}                              & \textbf{Value}         \\ \midrule
    \multirow{14}{*}{CARLA} & \multirow{10}{*}{Non-diffusion} & Epochs (One-stage training) & 30 \\
                       &                            & Epochs (Two-stage training)                     & 30                     \\
                       &                            & Loss weight: diffusion                    & 1.0                    \\
                       &                            & Loss weights: semantic, BEV, depth, speed       & 1.0                    \\
                       &                            & Loss weights: center heatmap, size, offset, yaw & 1.0                    \\
                       &                            & Prediction horizon (Timesteps)                  & 10                     \\
                       &                            & Learning rate                                   & 3e-4                   \\
                       &                            & Batch size (One-stage training)                 & 256                    \\
                       &                            & Batch size (Two-stage training)                 & 16                     \\
                       &                            & Weight decay                                    & 0.01                   \\ \cmidrule(l){2-4} 
                       & \multirow{4}{*}{Diffusion} & Noise schedule                                  & Square Cosine Schedule \\
                       &                            & Noise coefficient                               & 0.0001, 0.02           \\
                       &                            & Number of forward diffusion steps               & 100                    \\
                       &                            & Number of reverse denoising steps               & 2                      \\ \bottomrule
    \end{tabular}%
    }
    \end{table}

\section{Details of Experimental Setup}\label{app:experiment_setup}
\subsection{CARLA Benchmarks}\label{app:experiment_carla}
\subsubsection{Implementation Details}
This research is based on a dataset constructed from the MPC expert in the TF++ framework \cite{jaeger2023hidden}, containing 750k frames of high-quality driving data. We employ the RegnetY-3.2GF \cite{radosavovic2020designing} network as the encoder for image and LiDAR inputs. To optimize computational resources and improve training efficiency, we designed a two-stage training strategy: the first stage focuses on the perception module, utilizing multi-task loss functions including semantic segmentation, depth estimation, object detection bounding boxes, and bird's-eye view prediction; the second stage freezes the trained perception module and uses its output as conditional information to train the diffusion decoder. The entire training process consists of 30 epochs for each stage, with an initial learning rate of 3e-4, and optimized batch sizes for different stages—16 for the first stage and 256 for the second stage to accelerate the convergence of the diffusion model. The specific hyperparameter settings are shown in Table \ref{tab:hyper_carla}. All experiments were conducted on four NVIDIA 3090 GPUs.

In the CARLA autonomous driving evaluation system, we use three core metrics to measure model performance. First is Route Completion (RC), which quantifies the percentage of the predetermined route completed by the vehicle, calculated as $\text{RC} = \frac{D_{\text{traveled}}}{D_{\text{total}}} \times 100\%$, where $D_{\text{traveled}}$ is the actual distance traveled and $D_{\text{total}}$ is the total route length. Second is the Infraction Score (IS), used to evaluate driving safety by accumulating penalty coefficients for violations: $\text{IS} = \prod_{i=1}^{N} p_i$, where $p_i$ is the penalty coefficient for the $i$-th violation (such as pedestrian collision 0.50, vehicle collision 0.60, static object collision 0.65, running red lights 0.70, running stop signs 0.80, etc.), and $N$ is the total number of violations. Finally, the Driving Score (DS) serves as a comprehensive evaluation metric, combining completion and safety: $\text{DS} = \text{RC} \times \text{IS}$. A higher DS value indicates stronger ability to complete tasks while ensuring safety, making it a key indicator for evaluating the overall performance of autonomous driving systems.

\subsubsection{Benchmarks}
\begin{table}[]
    \caption{Hyperparameters of \name{} on NAVSIM.}
    \label{tab:hyper_navsim}
    \resizebox{\textwidth}{!}{%
    \begin{tabular}{@{}l|l|lc@{}}
    \toprule
    \textbf{Benchmark}       & \textbf{Type}                   & \textbf{Parameter}                        & \textbf{Value}         \\ \midrule
    \multirow{14}{*}{NAVSIM} & \multirow{10}{*}{Non-diffusion} & Epochs                                    & 100                    \\
                             &                                 & Batch size                                & 64                     \\
                             &                                 & Prediction horizon (Timesteps)            & 8                      \\
                             &                                 & Learning rate                             & 1e-4                   \\
                             &                                 & Ego progress weight                       & 5.0                    \\
                             &                                 & Time to collision weight                  & 5.0                    \\
                             &                                 & Comfort weight                            & 2.0                    \\
                             &                                 & Loss weights: diffusion                   & 10.0                   \\
                             &                                 & Loss weights: agent class, agent box, BEV & 10.0,1.0,10.0          \\
                             &                                 & Number of bounding boxes                  & 30                     \\ \cmidrule(l){2-4} 
                             & \multirow{4}{*}{Diffusion}      & Noise schedule                            & Square Cosine Schedule \\
                             &                                 & Noise coefficient                         & 0.0001, 0.02           \\
                             &                                 & Number of forward diffusion steps         & 100                    \\
                             &                                 & Number of reverse denoising steps         & 2                      \\ \bottomrule
    \end{tabular}%
    }
    \end{table}
\textbf{Longest6 Benchmark} is a high-difficulty, scalable evaluation benchmark designed to verify the performance of autonomous driving models in complex scenarios. Following \cite{chitta2022transfuser}, we use Longest6 as an alternative evaluation scheme for local ablation studies and multiple experiments. This benchmark selects the 6 longest routes from each town among the 76 training routes provided by CARLA, totaling 36 routes with an average length of 1.5 kilometers (close to the official leaderboard's 1.7 kilometers). To increase evaluation difficulty, we set the highest traffic density, including numerous dynamic obstacles (vehicles, pedestrians), combined with 6 weather conditions (such as Cloudy, Wet, HardRain) and 6 lighting conditions (such as Night, Dawn, Noon), while also including predefined adversarial scenarios (such as obstacle avoidance, unprotected left turns, pedestrians suddenly crossing, etc.). Compared to benchmarks like NoCrash and NEAT routes with shorter routes and lower traffic density, Longest6 more closely resembles the challenges of real scenarios, providing a more comprehensive validation of model robustness in complex urban environments (such as multi-lane intersections and dense traffic).

\textbf{Town05 Benchmark} is another important evaluation benchmark in CARLA, divided into Town05 Short and Town05 Long settings. Town05 Short includes 10 short routes (100-500 meters), each with 3 intersections; Town05 Long includes 10 long routes (1000-2000 meters), each with 10 intersections. These two settings are specifically designed to verify model driving capabilities in high-density dynamic traffic environments (including vehicles and pedestrians) and complex scenarios (such as unprotected intersection turns and randomly appearing pedestrians). Typically, models perform better on the Short benchmark, while performance decreases on the Long benchmark, reflecting the challenges of handling complex scenarios and long-term planning in long-distance driving. We use the same evaluation metrics as Longest6 (DS, RC, IS) to measure model performance on the Town05 benchmark.

\subsubsection{Baselines}\label{app:baselines_carla}
\begin{itemize}[leftmargin=*]
    \item \textit{WOR}~\cite{chen2021learning}: A model learning method based on the``world-on-rails" assumption, computing action value functions through offline dynamic programming to achieve driving policy distillation without actual interaction.
    \item \textit{LAV}~\cite{chen2022learning}: A full-vehicle learning framework that enhances training data diversity through viewpoint-independent BEV representation and trajectory prediction modules utilizing surrounding vehicle trajectories. The authors released two versions: LAV v1 and LAV v2.
    \item \textit{Interfuser}~\cite{shao2023safety}: A safety-enhanced multi-modal sensor fusion framework that integrates multi-view camera and LiDAR information through Transformers, outputting interpretable safety mental maps to constrain control actions.
    \item \textit{Transfuser}~\cite{chitta2022transfuser}: A Transformer-based multi-modal fusion architecture that fuses image and LiDAR features through cross-view attention mechanisms, enabling navigation in complex long-distance scenarios.
    \item \textit{TCP}~\cite{wu2022trajectory}: A trajectory-control joint prediction framework combining GRU temporal modules and trajectory-guided attention mechanisms, integrating dual-branch outputs through scene-adaptive fusion strategies.
    \item \textit{PlanT}~\cite{renz2022plant}: An interpretable planning Transformer based on privileged states (pose/velocity/route), achieving global traffic element reasoning through object-centered representation and self-attention mechanisms, with attention weights visualizing decision bases.
    \item \textit{DriveAdapter}~\cite{jia2023driveadapter}: A perception-planning decoupling framework that aligns student perception features with teacher planning features through learnable adapters, enhancing policy safety through action-guided mask learning. DriveAdapter+TCP indicates combining the output part with TCP\cite{wu2022trajectory}.
    \item \textit{ThinkTwice}~\cite{jia2023think}: A cascaded decoding paradigm that achieves action iterative optimization through a three-stage``observe-predict-refine" mechanism, utilizing scene feature retrieval and future state prediction.
    \item \textit{TF++}~\cite{jaeger2023hidden}: An improved version of Transfuser that eliminates target point following bias through a Transformer decoder. \textit{TF++} uses explicit uncertainty modeling with path+speed classification, while the \textit{TF++WP} variant retains waypoint prediction, achieving deceleration decisions through longitudinal averaging of continuous waypoints.
\end{itemize}

\subsubsection{Other Results}
In addition to CARLA Longest6, this research also rigorously evaluated the \name{} method on the CARLA Town05 Long and Town05 Short benchmarks, with detailed results shown in Table \ref{tab:town05_long} and Table \ref{tab:town05short}. On the Town05 Long benchmark, \name{} achieved a DS of 90.8 and an RC of 100, outperforming existing methods. On the Town05 Short benchmark, where environmental complexity is relatively lower, all algorithms generally showed improved performance, with the \name{} method still demonstrating excellent results, achieving a DS of 95.2 and an RC of 99.7. The outstanding performance across Town05 scenarios of varying complexity further validates the robustness and generalization capability of our method.
\begin{table}[]
    \centering
    \caption{\small {Comparison on the CARLA Town05 Long benchmark.}}
    \label{tab:town05_long}
    \resizebox{0.9\textwidth}{!}{%
    \begin{tabular}{@{}l|ccc|
    >{\columncolor[HTML]{CDE8F8}}c cc@{}}
    \toprule
    \textbf{Method}         & \textbf{Traj. Decoder} & \textbf{Img. Encoder} & \textbf{Input} & \textbf{DS $\uparrow$} & \textbf{RC $\uparrow$} & \textbf{IS $\uparrow$} \\ \midrule
    CILRS\cite{codevilla2019exploring}     & Explicit Policy & ResNet-34\cite{he2016deep}       & C    & 7.8        & 10.3       & 0.75       \\
    LBC\cite{chen2020learning}       & Explicit Policy & ResNet-18,34\cite{he2016deep}    & C    & 12.3       & 31.9       & 0.66       \\
    Roach\cite{zhang2021end}     & Explicit Policy & ResNet-34\cite{he2016deep}       & C    & 41.6       & 96.4       & 0.43       \\
    ST-P3\cite{hu2022st}     & Explicit Policy & EfficientNet-B4\cite{tan2019efficientnet} & C    & 11.5       & 83.2       & -          \\
    VAD\cite{jiang2023vad}       & Explicit Policy & ResNet-50\cite{he2016deep}       & C    & 30.3       & 75.2       & -          \\
    MILE\cite{hu2022model}      & Explicit Policy & ResNet-18\cite{he2016deep}       & C    & 61.1       & 97.4       & 0.63       \\
    DriveMLM\cite{wang2023drivemlm} & Explicit Policy & ViT-g/14\cite{zhai2022scaling}        & C\&L & 76.1       & 98.1       & 0.78       \\
    VADv2\cite{chen2024vadv2}     & Explicit Policy & ResNet-50\cite{he2016deep}       & C    & {\ul 85.1} & {\ul 98.4} & {\ul 0.87} \\ \midrule
    \textbf{DiffE2E (Ours)} & Diffusion Policy        & RegNetY-3.2GF\cite{radosavovic2020designing}         & C\&L           & \textbf{90.8}          & \textbf{100}           & \textbf{0.91}          \\ \bottomrule
    \end{tabular}%
    }
    \end{table}

\subsection{NAVSIM}\label{app:experiment_navsim}

\subsubsection{Implementation Details}
This research builds a model training framework based on NAVSIM's navtrain dataset. Unlike the image encoding architecture used in the CARLA simulation environment, we adopted VovNetV2-99\cite{lee2019energy} as the feature extraction backbone network in our NAVSIM experiments, a choice inspired by the Hydra-MDP\cite{li2024hydra} method, which can more effectively capture visual features in complex driving scenarios. All experiments were conducted on four NVIDIA 3090 GPUs.

The navtrain dataset establishes a highly complex driving environment evaluation system, characterized by carefully selected challenging scenarios with frequently changing driving intentions, deliberately excluding low-complexity driving situations such as stationary states and constant-speed cruising. This evaluation system quantitatively analyzes planning algorithm performance through a combination of non-reactive simulation and closed-loop evaluation mechanisms. This study uses the Predictive Driver Model Score (PDMS) as a comprehensive performance metric, which integrates performance across multiple key driving dimensions through weighted integration, including No at-fault Collision (NC), Drivable Area Compliance (DAC), Time-To-Collision (TTC), Comfort (C), and Ego Progress (EP). The detailed calculation methods for each metric are as follows:

The PDMS used in this study comprehensively evaluates autonomous driving policy performance by integrating sub-metrics across multiple key driving dimensions. Below is a detailed explanation of the definition and calculation method for each sub-metric:

\begin{itemize}[leftmargin=*]
    \item \textbf{No at-fault Collision}: Measures whether the ego vehicle has responsibility collisions with other traffic participants during simulation. The calculation adopts a tiered penalty mechanism: if a responsibility collision occurs (such as active collision in dynamic scenarios), then $\text{score}_{NC} = 0$, resulting in a PDMS of 0 for the entire scenario; if there is a collision with static objects (such as stationary vehicles), then $\text{score}_{NC} = 0.5$; if the ego vehicle is stationary or is rear-ended by a vehicle from behind (non-responsibility collision), no penalty is counted, $\text{score}_{NC} = 1$.
    
    \item \textbf{Drivable Area Compliance}: Evaluates whether the ego vehicle always remains within the drivable area. If the ego vehicle deviates from the drivable area, then $\text{score}_{DAC} = 0$, resulting in a PDMS of 0; if compliant throughout, then $\text{score}_{DAC} = 1$.
    
    \item \textbf{Time-To-Collision}: Detects whether the minimum collision time between the ego vehicle and other vehicles is below the safety threshold. Default value $\text{score}_{TTC} = 1$; if during the 4-second simulation, the predicted collision time between the ego vehicle and other vehicles is below the threshold (typically 2 seconds), then $\text{score}_{TTC} = 0$.
    
    \item \textbf{Comfort}: Measures trajectory smoothness, based on thresholds for acceleration and jerk. Points are deducted if acceleration or jerk exceeds predefined thresholds (such as acceleration $\leq$ 3.0 m/s², jerk $\leq$ 5.0 m/s³). The final score is calculated by comparing the deviation of actual parameters from thresholds, normalized to the $[0,1]$ interval.
    
    \item \textbf{Ego Progress}: Evaluates the efficiency of the ego vehicle's progress along the predetermined path, comparing actual progress with theoretical maximum safe progress. The calculated score $\text{score}_{EP}$ is the ratio of actual progress to theoretical maximum progress (limited between 0 and 1). Theoretical maximum progress is calculated by the PDM-Closed planner (based on search strategy for collision-free trajectories) as a safe upper limit; actual progress is the longitudinal movement distance of the ego vehicle along the path centerline during simulation. If the theoretical maximum progress is less than 5 meters, low or negative progress values are ignored.
\end{itemize}

\begin{table}[]
    \centering
    \caption{\small {Comparison on the CARLA Town05 Short benchmark.}}
    \label{tab:town05short}
    \resizebox{0.8\textwidth}{!}{%
    \begin{tabular}{@{}l|ccc|
    >{\columncolor[HTML]{CDE8F8}}c c@{}}
    \toprule
    \textbf{Method}         & \textbf{Traj. Decoder} & \textbf{Img. Encoder} & \textbf{Input} & \textbf{DS $\uparrow$} & \textbf{RC $\uparrow$} \\ \midrule
    CILRS\cite{codevilla2019exploring}      & Explicit Policy & ResNet-34\cite{he2016deep}       & C    & 7.5        & 13.4       \\
    LBC\cite{chen2020learning}        & Explicit Policy & ResNet-18,34\cite{he2016deep}    & C    & 31.0       & 55.0       \\
    Transfuser\cite{chitta2022transfuser} & Explicit Policy & RegNetY-3.2GF\cite{radosavovic2020designing}   & C\&L & 54.5       & 78.4       \\
    ST-P3\cite{hu2022st}      & Explicit Policy & EfficientNet-B4\cite{tan2019efficientnet} & C    & 55.1       & 86.7       \\
    NEAT\cite{chitta2021neat}       & Explicit Policy & ResNet-34\cite{he2016deep}       & C    & 58.7       & 77.3       \\
    Roach\cite{zhang2021end}      & Explicit Policy & ResNet-34\cite{he2016deep}       & C    & 65.3       & 88.2       \\
    WOR\cite{chen2021learning}        & -               & ResNet-18,34\cite{he2016deep}    & C    & 64.8       & 87.5       \\
    VAD\cite{jiang2023vad}        & Explicit Policy & ResNet-50\cite{he2016deep}       & C    & 64.3       & 87.3       \\
    VADv2\cite{chen2024vadv2}      & Explicit Policy & ResNet-50\cite{he2016deep}       & C    & 89.7       & 93.0       \\
    Interfuser\cite{shao2023safety} & Explicit Policy & ResNet-50\cite{he2016deep}       & C\&L & {\ul 95.0} & {\ul 95.2} \\ \midrule
    \textbf{DiffE2E (Ours)} & Diffusion Policy        & RegNetY-3.2GF\cite{radosavovic2020designing}         & C\&L           & \textbf{95.2}          & \textbf{99.7}          \\ \bottomrule
    \end{tabular}%
    }
\end{table}

PDMS integrates these sub-metrics through the following formula:
\begin{equation}
    \text{PDMS} = \underbrace{\left(\prod_{m\in\{NC, DAC\}}\text{score}_{m}\right)}_{\text{penalty term}} \times \underbrace{\left(\frac{5\cdot\text{score}_{EP} + 5\cdot\text{score}_{TTC} + 2\cdot\text{score}_{C}}{5 + 5 + 2}\right)}_{\text{weighted average term}}
\end{equation}

The key logic of this metric design includes: (1) Safety first: NC and DAC serve as hard penalty terms, ensuring policy must meet basic safety requirements; (2) Multi-dimensional balance: coordinating driving efficiency, safety, and comfort through weight allocation (EP and TTC have the highest weights, followed by C); (3) Scenario adaptability: filtering simple scenarios (such as stationary or straight driving), focusing on challenging conditions (such as turning, interaction), avoiding metric domination by trivial samples. This metric design aims to address the limitations of traditional displacement error (ADE), providing a simulation-driven multi-dimensional assessment that better reflects actual driving performance.

\subsubsection{Baselines}\label{app:baselines_navsim}
\begin{itemize}[leftmargin=*] 
    \item \textit{AD-MLP}~\cite{zhai2023rethinking}: A lightweight multi-layer perceptron (MLP) network that uses only vehicle state as input, without a perception module, performing trajectory prediction only.
    \item \textit{VADv2}~\cite{chen2024vadv2}: A probability-based planning framework that constructs a vocabulary of 4096 candidate trajectories through trajectory space discretization.
    \item \textit{UniAD}~\cite{hu2023planning}: The first multi-task end-to-end framework that unifies perception, prediction, and planning tasks, using a query mechanism to achieve module collaboration.
    \item \textit{LTF}~\cite{chitta2022transfuser}: A pure vision-based end-to-end driving framework that achieves implicit BEV representation through cross-modal attention mechanisms using images, replacing LiDAR input.
    \item \textit{PARA-Drive}~\cite{weng2024drive}: A fully parallel architecture that simultaneously executes perception, prediction, and planning tasks, improving inference speed by 3 times.
    \item \textit{Transfuser}~\cite{chitta2022transfuser}: A multi-modal end-to-end framework that fuses camera and LiDAR features, combining cross-modal attention and multi-scale feature interaction.
    \item \textit{LAW}~\cite{li2024enhancing}: A latent world model that enhances spatiotemporal feature learning through future state prediction, jointly optimizing scene understanding and trajectory planning.
    \item \textit{DRAMA}~\cite{yuan2024drama}: A Mamba-based planner combining multi-scale convolutional encoders and state Dropout, using an efficient Mamba-Transformer decoder to generate long trajectory sequences.
    \item \textit{Hydra-MDP}~\cite{li2024hydra}: A multi-teacher knowledge distillation framework that integrates human demonstrations and rule-based experts (such as traffic lights, lane keeping).
    \item \textit{DiffusionDrive}~\cite{liao2024diffusiondrive}: A truncated diffusion model for real-time sampling, reducing denoising steps to 2 through anchor-guided noise initialization while maintaining multi-modal coverage.
    \item \textit{GoalFlow}~\cite{xing2025goalflow}: A flow matching model with goal constraints, introducing a feasible region scoring mechanism and shadow trajectory refinement strategy.
    \item \textit{Hydra-MDP++}~\cite{li2025hydra}: An enhanced version of Hydra-MDP, adopting a V2-99\cite{lee2019energy} image encoder and expanding metric supervision (such as traffic lights, lane keeping, comfort).
\end{itemize}

\section{Visualization of Generated Trajectories}\label{app:visualization}
\begin{figure}
    \centering
    \includegraphics[width=\textwidth]{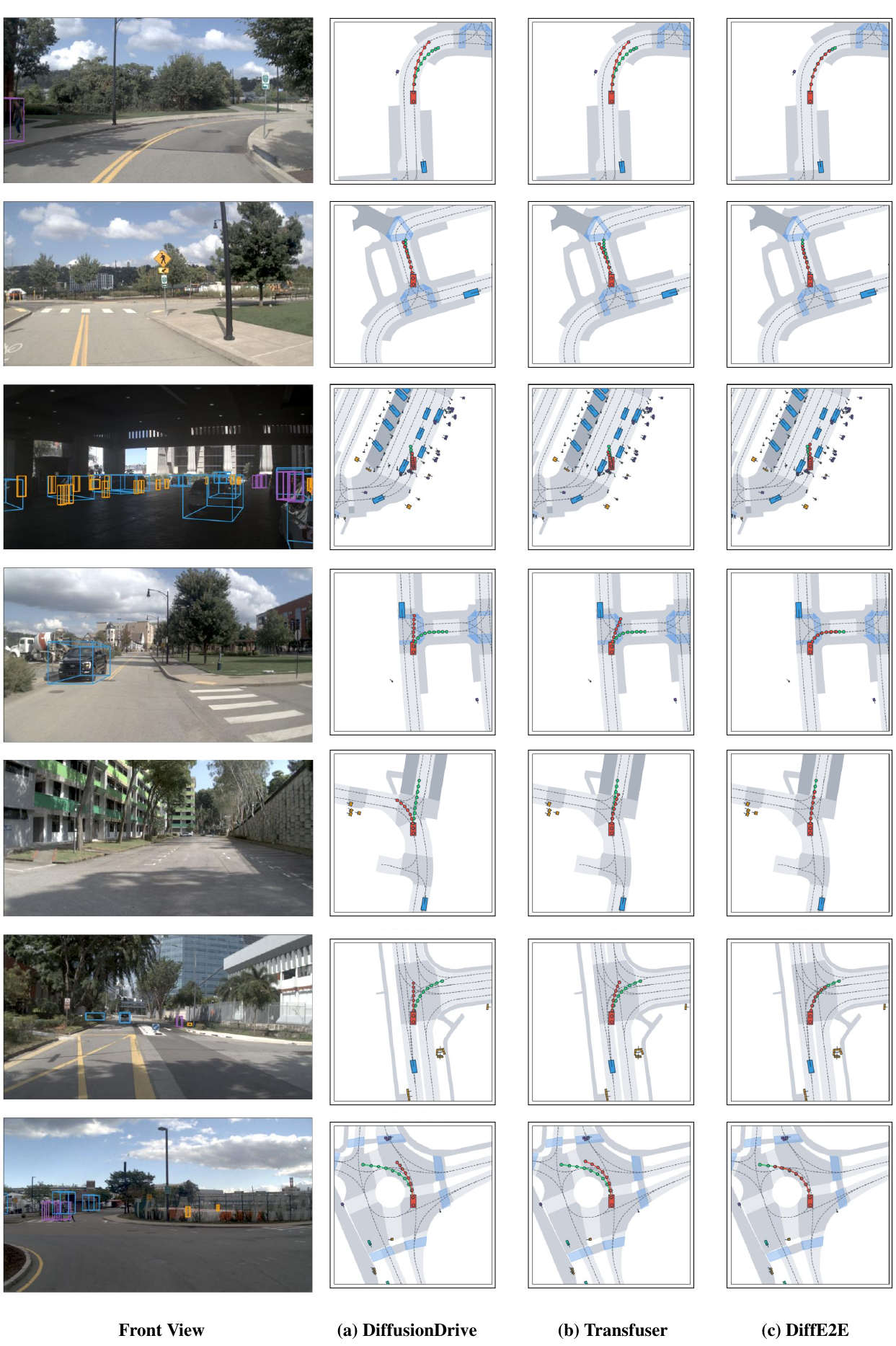}
    \vspace{-5pt}
    \caption{Visualization on Navtest benchmark.}
    \label{fig:appendix_navsim_1}
    \vspace{-5pt}
\end{figure}

\begin{figure}
    \centering
    \includegraphics[width=\textwidth]{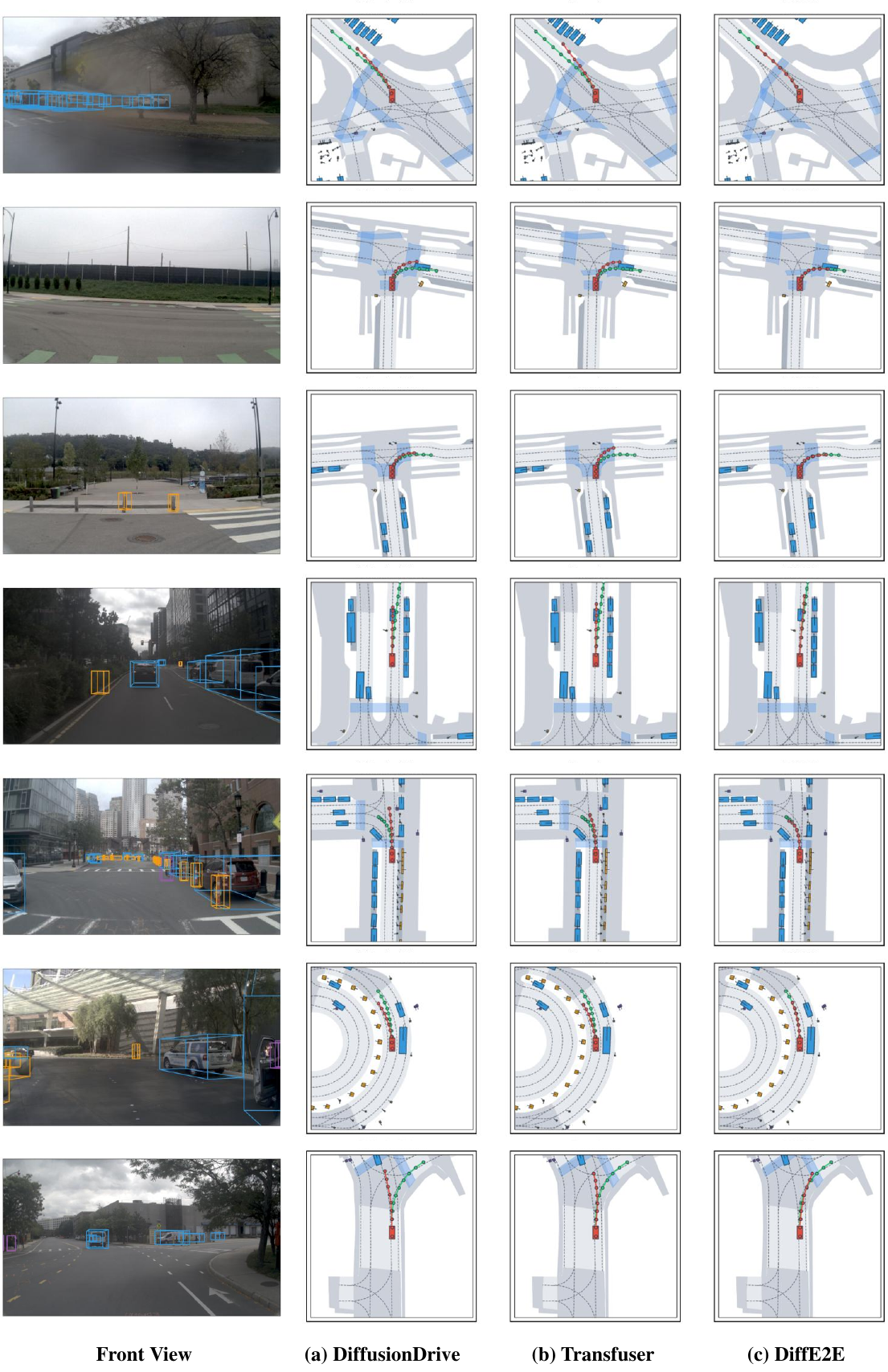}
    \vspace{-5pt}
    \caption{Visualization on Navtest benchmark.}
    \label{fig:appendix_navsim_2}
    \vspace{-5pt}
\end{figure}

We conducted a systematic comparative analysis of the trajectory generation capabilities of \name{}, DiffusionDrive\cite{liao2024diffusiondrive}, and Transfuser\cite{dauner2024navsim} across diverse driving scenarios, with detailed results shown in Figure \ref{fig:appendix_navsim_1} and Figure \ref{fig:appendix_navsim_2}.

\newpage
\section{Additional Studies}\label{app:additional_studies}
In this section, we will delve into three key research questions: RQ1: Real-time performance considerations of DiffE2E; RQ2: Analysis of the impact mechanism of diffusion model output types; RQ3:Generalization Evaluation in Out-of-Distribution Scenarios. Through systematic analysis of these questions, we aim to comprehensively reveal the technical advantages and potential limitations of the DiffE2E framework in end-to-end autonomous driving systems.

\subsection{RQ1:Impact Mechanism Analysis of Diffusion Model Output Types}\label{app:RQ1}
Regarding model output forms, we conducted an in-depth study of two output strategies for diffusion models: predicting noise versus directly predicting trajectories (original samples), as shown in Table \ref{tab:ablation}. The results were surprising—when the model outputs noise, the driving score plummets to 20.1, a staggering 76\% decrease compared to outputting trajectories, with vehicles barely able to follow lanes in the simulation environment. This enormous difference reveals that directly modeling the trajectory space is more effective than the noise space for tasks requiring high precision, such as autonomous driving.

We attribute this mainly to two factors: First, the trajectory space is inherently highly structured and semantically rich, with clear temporal correlations, spatial continuity, and direct mapping to driving task objectives (safety, efficiency). This allows the model to more effectively capture inherent patterns of driving behavior (such as car following, lane changing, turning, and other typical driving paradigms); in contrast, the noise space lacks clear physical meaning and semantic structure, requiring the model to learn complex mapping relationships from abstract noise to specific trajectories. Second, in the noise prediction paradigm, there is a significant error accumulation effect in the reverse generation process of diffusion models ($\bm{x}^{(T)} \rightarrow \bm{x}^{(T-1)} \rightarrow ... \rightarrow \bm{x}^{(0)}$)—small deviations in each denoising step are transmitted and amplified to subsequent steps, and as iterations progress, this cumulative error can cause the finally generated trajectory to severely deviate from the feasible driving space, which is particularly fatal in tasks requiring high precision like autonomous driving.

\subsection{RQ2:Real-Time Performance Considerations of DiffE2E}\label{app:RQ2}
\begin{wraptable}{r}{0.5\textwidth}
    \vspace{-5mm}
    \caption{\small Comparison of inference latency and model parameters between different methods.}
    \label{tab:time}
    \centering
    \resizebox{0.48\textwidth}{!}{%
    \begin{tabular}{@{}l
        >{\columncolor[HTML]{CDE8F8}}c cc@{}}
        \toprule
        Method         & Latency (ms) $\downarrow$ & Params & Img. Enc. \\ \midrule
        Transfuser\cite{dauner2024navsim}     & 29.1         & 55M    & ResNet-34\cite{he2016deep} \\
        DiffusionDrive\cite{liao2024diffusiondrive} & 39.1         & 60M   & ResNet-34\cite{he2016deep} \\
        GoalFlow\cite{xing2025goalflow}       & 66.8         & 190M   & V2-99\cite{lee2019energy}     \\
        \textbf{DiffE2E (Ours)}        & {42.8}         & {105M}    & V2-99\cite{lee2019energy}     \\ \bottomrule
        \end{tabular}%
    }
    \vspace{-2mm}
\end{wraptable}
In end-to-end autonomous driving systems, computational efficiency and real-time performance are crucial evaluation metrics. Although diffusion models have strong generative capabilities, they also suffer from high computational complexity. However, through our phased training strategy (separate optimization of perception module and diffusion decoder), we significantly improved the model's learning efficiency for trajectory distributions, enabling high-quality trajectory predictions with only two denoising iterations during inference. We analyzed the computational latency performance of \name{} compared to existing diffusion-based methods DiffusionDrive and GoalFlow, as well as the non-diffusion method Transfuser on the NAVSIM benchmark using a single NVIDIA RTX 3090 GPU, with results shown in Table \ref{tab:time}. The experimental data shows that Transfuser, based on the lightweight ResNet-34 image encoder, demonstrates the lowest latency (29.1ms), followed closely by DiffusionDrive (39.1ms); while \name{} and GoalFlow, which adopt the large VoVNetV2-99 encoder architecture, require 42.8ms and 66.8ms inference time respectively. Notably, latency performance shows a clear positive correlation with model parameter scale, reflecting the inherent trade-off between model complexity and computational efficiency. Nevertheless, \name{} achieves real-time performance comparable to lighter models while maintaining a relatively high parameter count (60M) through optimized diffusion sampling strategies, fully validating \name{}'s balanced optimization between computational efficiency and model expressiveness.

\subsection{RQ3:Generalization Evaluation in Out-of-Distribution Scenarios}\label{app:RQ3}
\name{} not only performs excellently on the training set but also demonstrates outstanding generalization ability in out-of-distribution test scenarios across both CARLA simulator and NAVSIM benchmark platforms. Particularly in complex traffic scenarios shown in Figure \ref{fig:carla}, \name{} exhibits highly intelligent contextual adaptation—when an obstacle vehicle blocks the regular right-turn route in the right lane, the model can dynamically adjust its decision, temporarily planning a safe straight path to avoid the vehicle ahead, and then completing the turning operation. This precise response and flexible adjustment to unseen traffic conditions fully demonstrates \name{}'s powerful generalization ability and decision intelligence when facing complex and variable driving environments.

This excellent generalization performance stems from the inherent mechanism advantages of diffusion models. During training, \name{} effectively builds robust mapping relationships from "non-ideal states" to target trajectories through systematic noise perturbation modeling and gradual denoising learning, enabling the model to handle various perception anomalies, rare behaviors, and unseen scenarios. More importantly, unlike traditional methods, \name{} models the complete trajectory probability distribution rather than a single deterministic solution. This distribution-based modeling approach naturally accommodates the inherent multi-modality and uncertainty in autonomous driving tasks, laying a solid foundation for the model's generalization deployment in complex real-world environments.

\section{Limitations \& Future Works}\label{app:limitations}
In this section, we discuss the main limitations and future work directions:
\begin{itemize}[leftmargin=*] 
    \item \textbf{Sampling Algorithm}. The current DDIM~\cite{song2020denoising} sampling algorithm faces computational efficiency bottlenecks in autonomous driving trajectory generation. Although \name{} can generate trajectories with just 2 denoising steps, more complex driving scenarios may require higher denoising costs to produce high-quality and diverse trajectories. Additionally, trajectory generation in complex traffic scenarios may lack sufficient safety and reliability guidance.

    \textit{Future work:} We plan to explore more efficient sampling algorithms such as DPM-Solver\cite{lu2022dpm} and Elucidated Diffusion Models (EDM)\cite{Karras2022edm}, which can significantly reduce sampling steps while maintaining generation quality through more precise ODE/SDE numerical approximations. We can also incorporate consistency models\cite{song2023consistency}, distillation algorithms\cite{meng2023distillation}, or Shortcut Models\cite{frans2024one} to accelerate sampling efficiency.
    Furthermore, we will employ classifier or energy function guidance techniques to constrain the sampling process by integrating prior knowledge of traffic rules and collision avoidance, thereby enhancing the safety and controllability of generated trajectories. Combining efficient sampling algorithms with domain-specific guidance mechanisms promises to significantly improve real-time performance while ensuring trajectory quality.
    
\end{itemize}

\end{document}